\documentclass[acmsmall,screen]{acmart}

\usepackage{bm}
\usepackage{xcolor}
\usepackage{textcomp}
\usepackage{stfloats}
\usepackage{verbatim}
\usepackage{multirow}
\usepackage{array}
\usepackage{pifont}
\usepackage{xspace}
\usepackage{dirtytalk}
\usepackage{xspace}
\usepackage{ragged2e}
\usepackage{tabularx}
\usepackage{makecell}
\usepackage{caption}
\usepackage{adjustbox}
\usepackage{booktabs}
\usepackage{threeparttable}

\newcommand{\meth}[4]{%
\textbf{#1}~{\footnotesize #2}\par
#3\par
{\footnotesize\(\displaystyle #4\)}%
}

\newcommand{\eg}{e.g.,\xspace}
\newcommand{\cmark}{\ding{51}}
\DeclareUnicodeCharacter{030C}{\v{ }}
\newcolumntype{L}[1]{>{\RaggedRight\arraybackslash}p{#1}}
\newcolumntype{C}[1]{>{\Centering\arraybackslash}p{#1}}
\newcolumntype{Y}{>{\RaggedRight\arraybackslash}X}
\newcolumntype{M}[1]{>{\centering\arraybackslash}m{#1}}

\usepackage{booktabs}
\usepackage{threeparttable}

\usepackage[final,markup=underlined,commandnameprefix=ifneeded]{changes}

\definechangesauthor[name={Kenny}, color=blue]{kenny}

\newcommand{\moderatetable}{%
  \setlength{\tabcolsep}{4pt}%
  \renewcommand{\arraystretch}{1.02}%
}

\newcolumntype{Z}{>{\Centering\arraybackslash}X} 

\AtBeginDocument{%
  }

\title{A Survey of Robotic Navigation and Manipulation with Physics Simulators in the Era of Embodied AI}

\author{Wong Lik Hang Kenny}
\email{klhwong3-c@my.cityu.edu.hk}
\orcid{0009-0004-3599-1649}
\affiliation{
  \institution{Department of Computer Science, City University of Hong Kong}
  \city{Hong Kong}
  \country{China}
}

\author{Xueyang Kang}
\authornote{Xueyang Kang is the corresponding author.}
\email{xueyang.kang@ntu.edu.sg}
\orcid{0000-0001-7159-676X}
\affiliation{
  \institution{School of Electrical and Electronic Engineering (EEE), Nanyang Technological University}
  \city{Singapore}
  \country{Singapore}
}

\author{Kaixin Bai}
\email{kaixin.bai@studium.uni-hamburg.de}
\orcid{0000-0001-8579-0547}
\affiliation{
  \institution{Department of Informatics, Universität Hamburg}
  \city{Hamburg}
  \country{Germany}
}
\affiliation{
  \institution{Agile Robots}
  \city{Munich}
  \country{Germany}
}

\author{Jianwei Zhang}
\email{jianwei.zhang@uni-hamburg.de}
\orcid{0000-0002-7856-5760}
\affiliation{
  \institution{Department of Informatics, Universität Hamburg}
  \city{Hamburg}
  \country{Germany}
}
\begin{document}
\settopmatter{authorsperrow=4}

\renewcommand{\shortauthors}
{Wong et al.}

\begin{abstract}
Navigation and manipulation are core capabilities in Embodied AI, but training agents to perform them directly in the real world is costly, time-consuming, and unsafe. Therefore, sim-to-real transfer has emerged as a key approach, yet the sim-to-real gap persists. This survey examines how physics simulators address this gap by analyzing properties that have received limited attention in prior surveys. We also analyze their features for navigation and manipulation tasks, as well as their hardware requirements. Additionally, we offer a resource with benchmark datasets, metrics, simulation platforms, and methods to help researchers select suitable tools while accounting for hardware constraints.
\end{abstract} 



\maketitle

\section{Introduction}

Navigation and manipulation are central capabilities of Embodied AI (EAI) for most modern robotic applications. Navigation allows an agent to move safely through the environment, while manipulation allows it to interact with objects to complete tasks. Recent advances in Artificial Intelligence have made learning-based methods---such as Reinforcement Learning (RL) and Imitation Learning (IL)---highly promising for training navigation and manipulation agents. However, collecting real-world interaction data to train these agents remains expensive, time-consuming, and sometimes unsafe. Physics-based simulation is therefore a central tool for robotic navigation and manipulation research as it supports scalable data generation, repeatable evaluation, controller prototyping, system debugging, and safer testing before hardware deployment \cite{liu2021physics_simulators}.

A robotics simulator typically comprises two levels. At the lower level, a \emph{physics engine} or \emph{dynamics engine} computes how the state of robots, objects, contacts, and environments evolves over time. At the higher level, a \emph{simulation platform} integrates the physics engine with scene construction, rendering, sensor simulation, robot models, scripting interfaces, benchmark tasks, and software bridges to robotics middleware (\eg Robot Operating System (ROS)). Representative physics engines include MuJoCo, Bullet, PhysX, DART, ODE, and Chrono, whereas representative simulation platforms include Gazebo, Isaac Sim, Habitat, AirSim, iGibson, Webots, and CoppeliaSim \cite{liu2021physics_simulators}. These engines differ not only in speed and rendering quality, but also in the physical phenomena they expose or approximate, such as rigid-body dynamics, contact and friction, actuator behavior, deformable materials, terrain response, fluids, and sensor noise.

The choice of physics engines and simulators depends strongly on the physical complexity required by the target task. For example, rigid-body collision and camera rendering may be sufficient for wheeled indoor navigation. For underwater navigation, however, hydrodynamics, degraded visibility, and specialized sensing become essential. For manipulation, especially contact-rich or deformable-object manipulation, accurate contact, friction, actuation, and sometimes tactile or deformable-body simulation become critical. Therefore, physics engines and simulators constrain what policies can be learned, what failures can be reproduced, and how much of the real-world behavior can be anticipated before deployment. Current research in navigation and manipulation increasingly relies on simulation experiments, yet there is still a lack of systematic understanding of how simulator properties (e.g., contact solvers, collision response, actuation timing, integration settings, sensor models, and rendering) affect learning, evaluation, and real-world deployment. As a result, two methods may appear comparable in simulation while transferring differently to hardware, or a policy may overfit to a simulator artifact that is absent in the physical world.

Current simulation-based evaluation in embodied AI is often not sufficiently predictive of real-world deployment. The challenge is to identify which simulator assumptions preserve the task-relevant behavior and failure modes after transfer. Evidence from embodied navigation shows that agents can exploit collision-response artifacts, such as wall sliding, leading to low sim-to-real correlation \cite{kadian2020sim2real}. On the other hand, evidence from manipulation shows that contact solvers, friction models, collision geometry, and actuation interfaces can substantially change object motion and grasp stability \cite{collins2019quantifying,kataoka2023bimanual}. Similar gaps appear in locomotion through latency, actuator, and foot-ground contact mismatch \cite{tan2018simtoreal}, in visual policies through lighting, texture, distractor, and camera perturbations \cite{chattopadhyay2021robustnav,pumacay2024colosseum}, and in deformable, tactile, and underwater settings where material, contact, optical, acoustic, and hydrodynamic effects remain difficult to model at scale \cite{blanco2024benchmarking,narang2021tactile,amer2023unavsim,oceansim2025}. Therefore, it is important to identify which properties of the simulator must be modeled, calibrated, randomized, or explicitly tested for a given navigation or manipulation task.

Motivated by this, this survey reviews navigation and manipulation, as well as the physics simulators used to train and evaluate agents. We examine how simulator properties interact with task requirements, representation learning, policy learning, and sim-to-real transfer. We also discuss the compute and hardware implications of simulator choices. Specifically, we ask three practical questions. \textbf{(1)} Which simulator properties matter most for different tasks in embodied navigation and manipulation? \textbf{(2)} Which sim-to-real gaps are captured, mitigated, or left unresolved by different simulator designs? \textbf{(3)} How should researchers select methods and configure simulators for modern navigation and manipulation tasks? Our scope is summarized in Figure \ref{fig:task_taxo}. Since our manipulation benchmark comparison focuses on success rate (SR) under task-specific success criteria, we do not include a separate metrics subsection. Moreover, instead of creating a separate tasks subsection, navigation tasks are categorized
as goal-driven or task-driven in the overview of
Section~\ref{sec:nav} and further detailed alongside their
benchmarks in Section~\ref{sec:nav_datasets}.

\begin{figure*}[ht]
    \centering
    \includegraphics[width=0.9\linewidth]{figures/taxonomy.pdf}
    \caption{Survey taxonomy covering tasks, simulators, datasets, metrics, and methods.}
    \Description{A taxonomy diagram organizing the survey around navigation and manipulation tasks, simulators, datasets, metrics, and methods.}
    \label{fig:task_taxo}
\end{figure*}

\vspace{-4mm}

\subsection{Sim-to-Real Transfer and Sim-to-Real Gap}

Sim-to-real transfer refers to deploying policies trained in simulation onto hardware. As simulated sensing, actuation, dynamics, and task conditions only approximate reality, policies that perform well in simulation can degrade on hardware; this discrepancy is the sim-to-real gap. For a policy \(\pi\) and a higher-is-better task metric \(\psi\), we can define the sim-to-real gap as \(G(\pi)=\psi_{\mathrm{sim}}(\pi)-\psi_{\mathrm{real}}(\pi)\), where \(\psi_{\mathrm{sim}}\) and \(\psi_{\mathrm{real}}\) evaluate the same policy in simulation and reality. Beyond absolute performance, a simulator should also be predictive: methods that rank highly in simulation should rank highly on hardware, as measured, for example, by the Sim-to-Real Correlation Coefficient (SRCC) \cite{kadian2020sim2real}.

In this survey, we group sim-to-real discrepancies into two broad categories: the perception gap and the action-dynamics gap. The perception gap covers differences in sensory observations, including RGB-D images, LiDAR, lighting, textures, calibration, and noise. The action-dynamics gap covers differences in robot--environment interaction, including collision response, contact, friction, actuation, controller timing, deformable dynamics, terrain response, fluids, and proprioceptive or tactile feedback. Common mitigation strategies can be grouped into three categories. First, \emph{calibration and explicit modeling} methods try to reduce known discrepancies by making the simulator closer to the target system, for example, through sensor-noise modeling, latency modeling, actuator calibration, and system identification of physical parameters such as mass, friction, damping, or controller gains. Second, \emph{randomization-based methods} deliberately train policies over a distribution of simulated conditions rather than a single nominal simulator; this includes visual domain randomization for textures, lighting, camera parameters, and observation noise, as well as dynamics randomization for contact, friction, payload, delays, and actuator properties \cite{tobin2017domain,peng2018sim,openaidex,akkaya2019solving,ma2024dreureka}. Third, \emph{transfer-oriented policy learning} aims to make the learned policy robust or adaptive to residual mismatch, using techniques such as robust policy optimization, domain adaptation, residual learning, or online adaptation during deployment \cite{chebotar2019closing}.

In parallel with the rapid development of analytical physics simulators, recent Embodied AI research has also shown growing interest in learned world models. We briefly discuss this trend here because world models represent another way of modeling environment dynamics and agent--environment interaction, and they have become increasingly relevant for planning, representation learning, and policy learning. A world model is typically a data-driven predictive model that uses past observations, actions, and sometimes latent memory to predict future observations, states, object poses, or latent representations. Recent studies and benchmarks suggest that physical consistency and physics-informed representations are important for world models, especially when they are used for prediction, planning, or manipulation-oriented policy learning \cite{pinwm2025,physworld2025}. This suggests that for embodied navigation and manipulation, the physical assumptions behind an environment model matter. Whether the environment is represented by an analytical simulator, a learned world model, or a hybrid of the two, properties such as contact behavior, actuation timing, sensing noise, deformable dynamics, and physical consistency can strongly affect learning, evaluation, and sim-to-real transfer. Table \ref{tab:sim2real_evidence} includes both analytical simulators and representative learned world models to illustrate that low-level modeling choices and physical assumptions can substantially influence sim-to-real predictability and transfer outcomes. However, unlike physics engines and simulation platforms, world models are usually learned from finite data and are often tied to specific task distributions, observation spaces, or interaction regimes. They do not generally provide the same level of user-controllable geometry, dynamics, material properties, sensor configuration, actuation models, or contact parameters as analytical simulators.

\begin{table*}[htbp]
\caption{Effects of low-level physics and sensing choices on sim-to-real transfer. Arrows indicate the preferred direction. SRCC$_{\mathrm{SPL}}$: sim-to-real correlation on SPL; CD: Chamfer distance; PADC: Physics-Aware Digital Cousins.}
\label{tab:sim2real_evidence}
\centering
\footnotesize
\setlength{\tabcolsep}{4pt}
\renewcommand{\arraystretch}{1.05}

\begin{tabularx}{\textwidth}{@{}
L{3.0cm}
L{3.0cm}
L{4.0cm}
Y@{}}
\toprule
\textbf{Work} & \textbf{Low-level factor} & \textbf{Result} & \textbf{Implication} \\
\midrule
\multicolumn{4}{@{}l@{}}{\textbf{I. Analytical simulators}}\\
\midrule

Kadian et al.~\cite{kadian2020sim2real}
& Collision response
& SRCC\(_{\text{SPL}}\) \(\uparrow\): \textbf{0.875} with tuned collision response vs 0.603 with default sliding
& Collision settings affect navigation predictivity. \\

Collins et al.~\cite{collins2019quantifying}
& Contact engine / solver
& Pose error \(\downarrow\): \textbf{45.46} with V-REP Newton vs 133.61 with V-REP Bullet 2.78
& Contact solver choice affects manipulation accuracy. \\

Kataoka et al.~\cite{kataoka2023bimanual}
& Actuation timing model
& Real assembly success \(\uparrow\): \textbf{50\%} with noise and interpolation vs 0\% without them
& Modeling timing effects can change real-world success. \\

\midrule
\multicolumn{4}{@{}l@{}}{\textbf{II. Learned world models}}\\
\midrule

PIN-WM~\cite{pinwm2025}
& Physics-informed world model
& Real push success \(\uparrow\): \textbf{75\%} with PIN-WM + PADC vs 30\% with RoboGSim \cite{li2024robogsim}
& Physical priors can improve transfer in constrained domains. \\

PhysWorld~\cite{physworld2025}
& Physics-aware demo video synthesis
& Real-video CD \(\downarrow\): \textbf{0.010} with PhysWorld vs 0.012 with PhysTwin \cite{jiang2025phystwin}
& Structured physical modeling improves prediction fidelity. \\

\bottomrule
\end{tabularx}
\end{table*}

\vspace{-4mm}

\subsection{Current Trends in Embodied AI}

\begin{figure*}[ht]
    \centering
    \includegraphics[width=\linewidth]{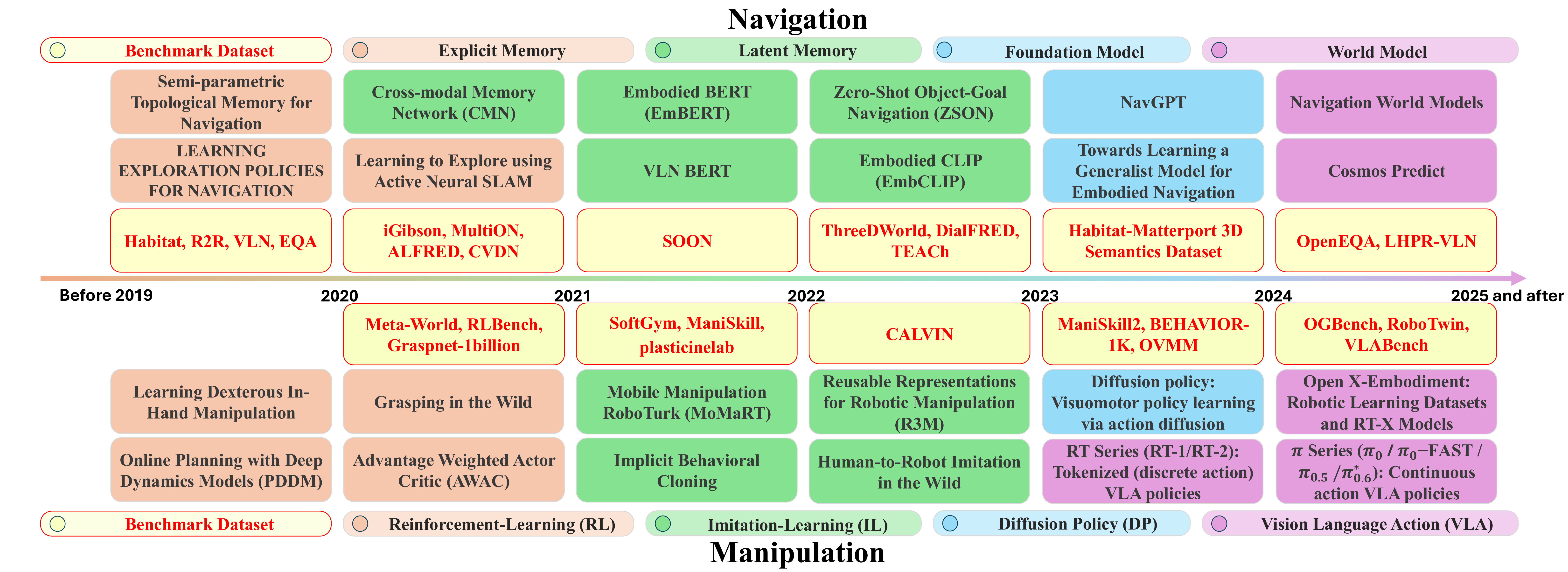}
    \caption{Evolution of navigation (top) and manipulation (bottom) methods and benchmarks since 2019, spanning memory, foundation/world models, RL, IL, diffusion policies, and VLAs.}
    \vspace{-3mm}
    \Description{A two-row timeline summarizing navigation and manipulation methods and benchmarks from 2019 onward.}
    \label{fig:timeline}
\end{figure*}

Figure \ref{fig:timeline} traces the shift toward large-scale data and model-centric learning. Navigation progressed from explicit and recurrent memory to latent memory \cite{Hong_2021_CVPR}, foundation-model planning \cite{zhou2023navgpt}, and world models \cite{bar2024navigationworldmodels,nvidia2025cosmosworldfoundationmodel}, alongside iGibson, ALFRED, and Habitat datasets \cite{li2021igibson20objectcentricsimulation,shridhar2020alfredbenchmarkinterpretinggrounded,yadav2023habitatmatterport3dsemanticsdataset}. Manipulation moved from model-free RL towards imitation learning \cite{openaidex,nair2021awacacceleratingonlinereinforcement,nagabandi2019deepdynamicsmodelslearning}, diffusion policies and VLAs \cite{chi2024diffusionpolicyvisuomotorpolicy,brohan2023rt2visionlanguageactionmodelstransfer,open_x_embodiment_rt_x_2023,black2024pi0visionlanguageactionflowmodel}, supported by large-scale manipulation datasets \cite{fang2020graspnet,mu2021maniskillgeneralizablemanipulationskill,corl2020softgym,open_x_embodiment_rt_x_2023}.

Earlier surveys on Embodied AI (EAI) simulators and navigation were conducted before the emergence of large language models (LLMs) and world models, and do not reflect recent breakthroughs \cite{duan2022surveyembodiedaisimulators}. Other surveys provide broad EAI overviews or focus on specific methodologies (\eg foundation models) without emphasizing a physics simulator-property perspective spanning both navigation and manipulation \cite{liu2024aligningcyberspacephysical,xu2024surveyroboticsfoundationmodels,zheng2025survey}. Most discussions of simulator choice are either qualitative or limited to simulation-only comparisons; they rarely address low-level simulator properties, such as contact/impact formulation, integration/solver settings, actuation timing, and sensor models, which affect sim-to-real outcomes. Compared with these works, this survey connects simulator properties to task requirements, learning methods, and sim-to-real behavior across navigation and manipulation. Given the growth of EAI, this paper presents a structured exploration of navigation, manipulation, and the physics simulators supporting these tasks.

\section{Navigation} \label{sec:nav} 

Navigation is an important capability of Embodied AI agents, enabling their deployment in diverse real-world applications, including autonomous vehicles \cite{gaia}, personal assistants \cite{pa}, and rescue robots \cite{hazard}. However, training these agents directly in the real world poses significant challenges, including high costs, time constraints, safety risks, environmental setup overhead, and difficulties in collecting large-scale training data. To overcome these, sim-to-real transfer has become a popular approach. However, successful sim-to-real transfer requires addressing two key challenges. The \textbf{perception gap} concerns whether RGB-D, LiDAR, sonar, IMU, and other simulated observations reproduce task-relevant appearance, geometry, calibration, noise, and latency. The \textbf{action-dynamics gap} concerns whether collision, terrain, actuation, and proprioceptive feedback support locomotion policies that remain reliable on hardware.

\begin{figure*}[ht]
    \centering
    \includegraphics[width=\linewidth]{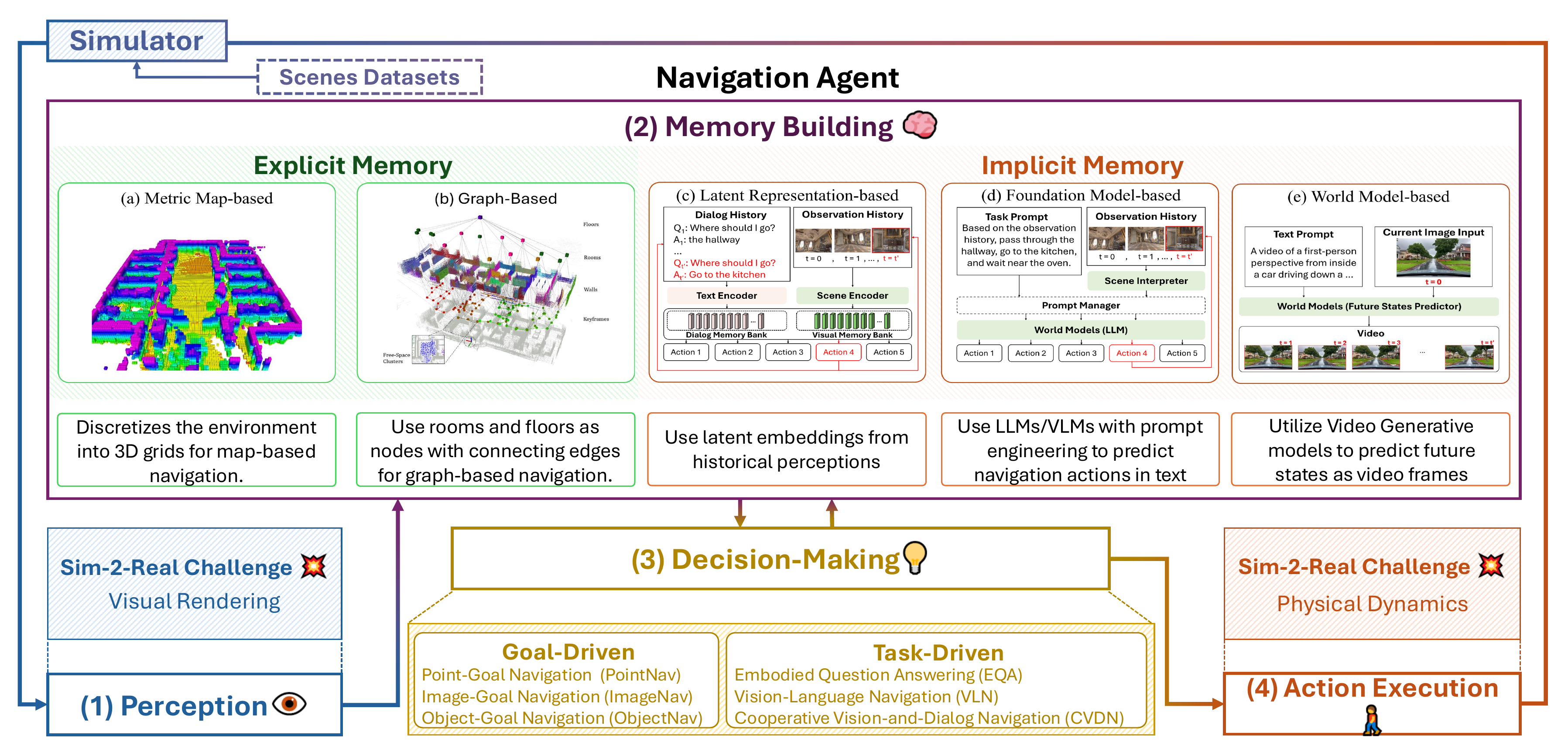}
\caption{Navigation pipeline: perception, explicit/implicit memory, decision-making, and execution for goal- and task-driven navigation, with perception and action-dynamics transfer gaps \cite{embodiedqa,mattersim}.}
\vspace{-3mm}
\Description{A navigation pipeline diagram showing perception, memory building, decision making, action execution, task categories, and sim-to-real gaps.}
\label{fig:nav_overview}
\end{figure*}

In Figure \ref{fig:nav_overview}, we present a breakdown of the navigation process in simulators, highlighting the four key steps and showing how the sim-to-real challenges emerge. The process begins with \textbf{(1) Perception}, where agents interpret sensory data from the simulator. This step is directly affected by the \textbf{perception sim-to-real gap}. Since agents typically have partial observability of their environment, they must engage in \textbf{(2) Memory Building}. This memory can be \textbf{Explicit}, using structures like metric maps \cite{metricmapnav} or topological graphs \cite{Rosinol20rss-dynamicSceneGraphs}, or \textbf{Implicit}, leveraging latent representations, foundation models (\eg LLMs, VLMs) \cite{zhou2023navgpt,zheng2023learning}, or world models \cite{bar2024navigationworldmodels}. With this memory, agents proceed to \textbf{(3) Decision-Making} for completing various navigation tasks. These tasks are broadly categorized as \textbf{goal-driven} (\eg PointNav, ObjectNav) and \textbf{task-driven}, where agents follow complex instructions for tasks like Embodied Question Answering (EQA) \cite{embodiedqa} or Vision-and-Language Navigation (VLN) \cite{mattersim}. Finally, in \textbf{(4) Action Execution}, the agent moves within the environment. This is where the \textbf{action-dynamics sim-to-real gap} becomes critical.

\subsection{Physics Engines and Simulators} \label{sec:nav_simulators}

Modern navigation simulators can be grouped into three categories based on their operational domain and the scalability of environments they support: \textbf{(1) Indoor simulators} are often tailored for structured, smaller-scale settings; \textbf{(2) Outdoor simulators} are designed for large-scale, dynamic environments; and \textbf{(3) General-purpose simulators} can be customized for multiple domains.

\begin{figure}[ht]
    \centering
    \includegraphics[width=\linewidth]{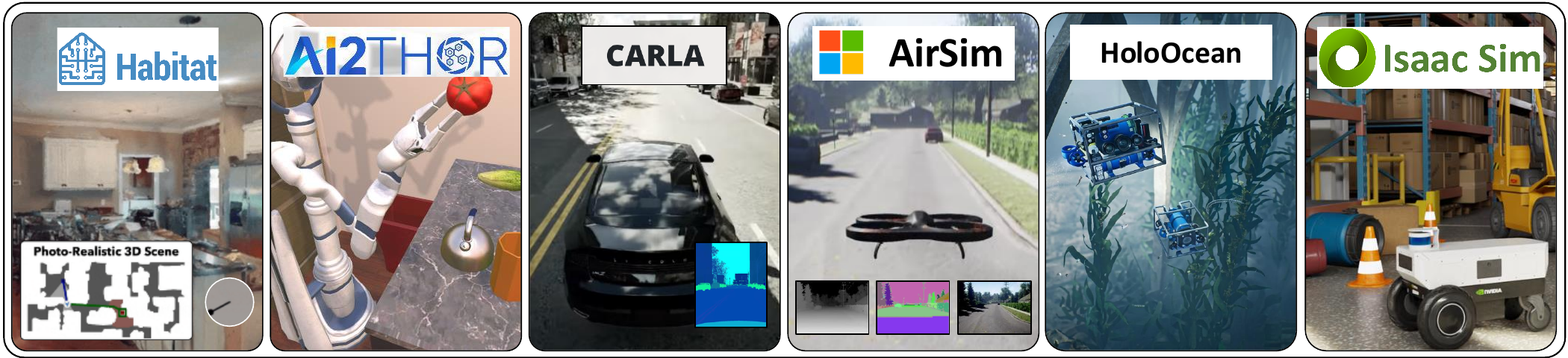}
    \caption{Navigation simulators across domains (left to right): Habitat, AI2-THOR, CARLA, AirSim, HoloOcean, and Isaac Sim.}
    \vspace{-3mm}
    \Description{Representative rendered scenes from Habitat, AI2-THOR, CARLA, AirSim, HoloOcean, and Isaac Sim.}
    \label{fig:nav_sim}
\end{figure}

Moreover, the scale of environments supported by these simulators can influence the selection of navigation tasks and the design of navigation agents. For example, explicit memory may suffice for smaller-scale indoor settings rendered by indoor simulators, whereas world models can help predict and represent long-horizon dynamics in large-scale environments provided by outdoor simulators. As discussed above, simulators must address two critical sim-to-real challenges--- the perception sim-to-real gap and the action-dynamics sim-to-real gap---to enable effective real-world deployment. Therefore, this section presents an in-depth analysis of these simulators, exploring their physical properties and how they mitigate these sim-to-real challenges. Figure \ref{fig:nav_sim} shows representative navigation simulators across indoor, outdoor, and general-purpose domains.

\noindent \textbf{(1) Indoor Simulators.} Indoor simulators cater to structured, small-scale environments (\eg homes) for applications such as personal assistants. Early works leveraged real-world environment scans or high-quality rendering to minimize the perception sim-to-real gap. For example, the \textbf{Matterport3D Simulator} \cite{Matterport3D} uses the Matterport3D dataset (10,800 panoramic views derived from 194,400 RGB-D images across 90 real-world scenes), achieving high visual fidelity. However, it lacks physical dynamics and restricts navigation to discrete transitions between precomputed viewpoints (averaging 2.25 meters apart) with implicit collision via walkable paths, making it unsuitable for learning locomotion over uneven terrain. By contrast, numerous simulators integrate physics engines to simulate interactive dynamics. For example, \textbf{Habitat-Sim} \cite{habitat19iccv, szot2021habitat, puig2023habitat3} employs the Bullet physics engine for rigid-body dynamics, thereby enabling collision-aware embodied motion and rigid-body interaction in 3D environments. It also renders reconstructed scenes from Matterport3D \cite{Matterport3D}, Gibson \cite{xiazamirhe2018gibsonenv}, HM3D \cite{ramakrishnan2021hm3d}, and synthetic CAD models. Additionally, it integrates RGB-D sensor noise models to simulate real-world imperfections, such as distortions, thereby improving sim-to-real transferability for perception models. Similarly, \textbf{AI2-THOR} \cite{kolve2022ai2thorinteractive3denvironment} provides photorealistic visuals via Physically-Based Rendering (PBR) from Unity3D and supports domain randomization (\eg varying materials and lighting) to improve sim-to-real transfer. It uses a Unity-based physics engine to enable realistic collision detection and locomotion. Meanwhile, \textbf{iGibson} \cite{xiazamirhe2018gibsonenv} uses PBR with Bidirectional Reflectance Distribution Function (BRDF) models to simulate realistic lighting and allows domain randomization to narrow the perception sim-to-real gap.

\noindent \textbf{(2) Outdoor simulators.} Outdoor simulators are designed for large-scale, dynamic environments. We categorize them into three subdomains---Ground/Wildland, Air, and Underwater---each of which imposes distinct requirements on both visual sensing and physical dynamics.

\noindent \textbf{(2a) Ground/Wildland.} \textbf{CARLA} \cite{Dosovitskiy17} offers simulated environments of urban areas and natural terrains for autonomous vehicles and robots. It harnesses Unreal Engine 4 to deliver photorealistic rendering through ray tracing and physically based rendering (PBR), while employing domain randomization to minimize the perception sim-to-real gap. On the action-dynamics side, it utilizes PhysX \cite{nvidia2020physx} to manage vehicle dynamics, collision detection, and locomotion. However, these advanced features entail significant computational overhead.

\noindent \textbf{(2b) Air.} \textbf{AirSim} \cite{airsim2017fsr} provides simulated environments for aerial robotics, such as drone flight. It similarly leverages Unreal Engine 4 for high-fidelity rendering and employs domain randomization to improve robustness against visual discrepancies. In modeling physical dynamics for action execution, AirSim uses a custom physics engine called Fast Physics, optimized for speed, to handle collision detection and integrates realistic sensor models, such as IMU and GPS, to accurately simulate real-world conditions.

\noindent \textbf{(2c) Underwater.} Underwater navigation requires specific physics modeling. Visually, light travels through water before reaching the camera housing, lens, and image sensor; during this process, refraction, wavelength-dependent absorption, and scattering change the observed image. For example, red wavelengths attenuate rapidly with depth, producing blue/green color shifts, while suspended particles, turbidity, marine snow, caustics, and low illumination reduce contrast and visibility \cite{gonzalez2023underwatercv}. Therefore, underwater simulators must model not only RGB cameras but also sensors such as imaging/side-scan sonar, Doppler Velocity Log (DVL), pressure/depth sensors, IMUs, and acoustic positioning or communication systems. Physically, underwater robots are strongly affected by hydrodynamic forces, including buoyancy, drag, lift, added mass, thruster dynamics, and ocean currents; these factors directly determine whether navigation and control policies transfer to real AUVs/ROVs. Consequently, underwater simulators must address both the perception sim-to-real gap from optical/acoustic sensing and the action-dynamics sim-to-real gap from fluid--robot interaction. \textbf{Stonefish} focuses on the physics side by explicitly modeling underwater hydrodynamics (\eg buoyancy, drag, and added mass) and providing marine sensor suites such as IMU/pressure sensors, DVL, sonar, and acoustic positioning/communication, making it suitable for dynamics-aware navigation and control \cite{cieslak2019stonefish}. \textbf{Project DAVE}, built in the ROS+Gazebo ecosystem, supports multi-phase underwater autonomy with parameterized ocean conditions such as currents and bathymetry, but its visual realism is limited by the rendering capability of Gazebo \cite{zhang2022dave}. In contrast, game-engine-based simulators such as \textbf{HoloOcean}, \textbf{MARUS}, and \textbf{UNav-Sim} leverage Unreal/Unity rendering and optical/sonar sensor synthesis to better approximate underwater observations, thereby helping reduce the perception sim-to-real gap while still providing vehicle dynamics for navigation \cite{potokar2022holoocean,lonvcar2022marus,amer2023unavsim}. More recently, \textbf{OceanSim} is built based on Isaac-Sim to provide high-fidelity and GPU-accelerated underwater sensor rendering \cite{oceansim2025}. It models underwater image formation with attenuation and backscatter, supports GPU-accelerated imaging sonar rendering, and provides DVL, IMU, and pressure/barometer sensor models. Furthermore, \textbf{MarineGym} is a recent platform that is also built on Isaac Sim. It supports GPU-accelerated hydrodynamics, Gym-compatible RL interface, domain randomization toolkit, multiple UUV models, and standardized control tasks, allowing scalable RL training for underwater robots \cite{marinegym2025}. Despite these advances, existing simulators still cannot fully reproduce the coupled complexity of real oceans: spatially and temporally varying turbidity, refractive camera-housing effects, acoustic multipath and noise, sediment disturbance, biofouling, deformable vegetation, turbulent currents, wave--vehicle interactions, and high-fidelity fluid--structure coupling remain difficult to simulate at scale. In practice, low-order 6-DoF hydrodynamic models may be sufficient for coarse waypoint navigation, whereas close-range docking, manipulation, near-seabed inspection, and miniature underwater robots require much higher fluid-dynamics fidelity that current simulators only approximate.

\noindent \textbf{(3) General-purpose simulators.} General-purpose simulators support both indoor and outdoor settings, delivering high-fidelity visuals and precise physics. However, these properties also entail a trade-off in computational complexity, often requiring high-end GPUs.
\textbf{ThreeDWorld (TDW)} \cite{gan2021threedworldplatforminteractivemultimodal}, built on Unity3D, uses PBR and High Dynamic Range Image (HDRI) lighting for photorealistic rendering, minimizing the perception sim-to-real gap. It employs the PhysX physics engine, which supports cloth and fluid simulations and ensures realistic collision detection. In addition, TDW supports tasks such as the Transport Challenge \cite{gan2021threedworldplatforminteractivemultimodal}, which require both indoor manipulation and outdoor exploration. Meanwhile, \textbf{Isaac Sim} \cite{isaacsim}, developed by NVIDIA, leverages RTX technology for photorealistic ray-traced rendering, thereby minimizing the perception sim-to-real gap. It also leverages the PhysX engine to provide precise dynamics for collision detection and locomotion. Integrated with Isaac Lab, it supports reinforcement learning and imitation learning for tasks like path planning, scaling from warehouses to outdoor settings. The high-fidelity simulations supported by these simulators ensure that policies trained are grounded in realistic visuals and physics, enhancing their deployability in real-world scenarios. However, to address the high computational demands, a hierarchical training framework is often used: a deep learning-based path planner is used for waypoint prediction, followed by an RL or Proportional-Derivative (PD) controller that manages locomotion and short-range navigation \cite{choi2024canvascommonsenseawarenavigationintuitive}.

\subsection{Benchmark Datasets and Environments} \label{sec:nav_datasets}

Navigation tasks in Embodied AI include goal-driven navigation, in which agents pursue a predefined target, and task-driven navigation, in which agents interpret and act based on complex textual instructions. To facilitate the training and evaluation of agents on these tasks in simulators, a variety of benchmark datasets have been proposed, each often paired with a specific simulation platform. This section offers a detailed examination of the navigation benchmark datasets. Table \ref{tab:nav_datasets} summarizes key properties of these benchmark datasets.

\begin{table}[t]
\caption{Benchmark datasets and environments for navigation tasks. Baseline success rate (SR) is reported when a directly comparable SR is available in the original paper; ``--'' indicates that SR is not reported or is not the primary metric, since some benchmarks use task-specific metrics, e.g., LLM-Match and efficiency for A-EQA, instance/localization success variants for ION, ISR/CSR/CGT for LHPR-VLN, and distance-based task scores or perception accuracy for OceanGym.}
\label{tab:nav_datasets}
\centering
\footnotesize
\moderatetable
\begin{threeparttable}
\begin{tabularx}{\columnwidth}{@{}
L{0.18\columnwidth}
L{0.24\columnwidth}
L{0.18\columnwidth}
Y
@{}}
\toprule
\multicolumn{4}{@{}l}{\textbf{Goal-Driven Navigation}}\\
\midrule
\textbf{Dataset} & \textbf{Size} & \textbf{SIM./SENS.} & \textbf{Baseline (SR)} \\
\midrule
iGibson~\cite{8954627,9636667,li2021igibson20objectcentricsimulation}
& 100+ SC.; 27K OBJ. DESC.
& iGibson / RGB-D-SEG
& -- \\

ION~\cite{10.1145/3474085.3475575}
& 600 SC.
& AI2-THOR / RGB
& -- \\

HM3D~\cite{ramakrishnan2021hm3d}
& 1,000 SC.
& Habitat / RGB-D
& DD-PPO~\cite{wijmansdd}: 97\% \\

HM3D-OVON~\cite{yokoyama2024hm3dovondatasetbenchmarkopenvocabulary}
& 181 SC.
& Habitat / RGB-D
& DAgRL+OD~\cite{yokoyama2024hm3dovondatasetbenchmarkopenvocabulary}: 37--39\% \\

MultiON~\cite{gireesh2023sequenceagnosticmultiobjectnavigation}
& 50K train EPS.
& Habitat / RGB-D
& OracleMap~\cite{gireesh2023sequenceagnosticmultiobjectnavigation}: 94\% (1-ON), 48\% (3-ON) \\

DivScene~\cite{wang2024divscenebenchmarkinglvlmsobject}
& 4,614 SC.; 81 types
& AI2-THOR / RGB
& NATVLM~\cite{wang2024divscenebenchmarkinglvlmsobject}: 54.9\% \\
\midrule
\multicolumn{4}{@{}l}{\textbf{Task-Driven Navigation}}\\
\midrule
\textbf{Dataset} & \textbf{Size} & \textbf{SIM./SENS.} & \textbf{Baseline (SR)} \\
\midrule
R2R~\cite{mattersim}
& 90 SC.; 21.6K INSTR.
& Matterport3D / RGB
& SEQ2SEQ~\cite{sutskever2014sequence}: 20.4\% \\

VLN-CE~\cite{krantz_vlnce_2020}
& 90 SC.; 4,475 TRAJ.
& Habitat / RGB-D
& Cross-modal attn.~\cite{krantz_vlnce_2020}: 33\% \\

VLN-CE-Isaac~\cite{cheng2024navila}
& 90 SC.; 1,077 TRAJ.
& Isaac Sim / RGB+LiDAR
& NaVILA~\cite{cheng2024navila}: 54.0\% (R2R) \\

ALFRED~\cite{shridhar2020alfredbenchmarkinterpretinggrounded}
& 120 SC.; 8K DEMOS; 25K DIR.; 428K pairs
& AI2-THOR / RGB
& SEQ2SEQ~\cite{sutskever2014sequence}: 4\% \\

DialFRED~\cite{Gao_2022}
& 112 rooms; 34.3K tasks
& AI2-THOR / RGB
& Q-P~\cite{Gao_2022}: 33.6\% \\

TEACh~\cite{Padmakumar_Thomason_Shrivastava_Lange_Narayan-Chen_Gella_Piramuthu_Tur_Hakkani-Tur_2022}
& 120 SC.; 3,047 SESS.
& AI2-THOR / RGB+SEG
& E.T.~\cite{pashevich2021episodic}: 9\% \\

VNLA~\cite{nguyen2019vision}
& 90 SC.; 94.8K tasks
& Matterport3D / RGB
& LEARNED~\cite{nguyen2019vision}: 35\% \\

REVERIE~\cite{qi2020reverie}
& 90 SC.; 21.7K INSTR.
& Matterport3D / RGB
& IN-P~\cite{qi2020reverie}: 11.28\% \\

A-EQA~\cite{OpenEQA2023}
& 180+ SC.; 1.6K+ Q
& Habitat / RGB-D
& -- \\

Robo-VLN~\cite{irshad2021hierarchical}
& 90 SC.; 3,177 TRAJ.
& Habitat / RGB-D
& HCM~\cite{irshad2021hierarchical}: 46\% \\

LHPR-VLN~\cite{song2024towards}
& 216 SC.; 3,260 tasks
& Habitat / RGB
& -- \\

OceanGym~\cite{oceangym2025}
& 800m\(\times\)800m env.; 8 DEC. tasks; 85 PERC. sets; 2 depths
& UE5.3+HoloOcean / 6-view RGB+Sonar
& -- \\

\bottomrule
\end{tabularx}
\begin{tablenotes}[flushleft]
\footnotesize
\item \textbf{Abbreviations:} SC.=scenes; EPS.=episodes; INSTR.=instructions; TRAJ.=trajectories; SESS.=sessions; DIR.=directives; Q=questions; SEG.=segmentation; RGB-D=RGB+Depth; SR=success rate; DEC.=decision; PERC.=perception; OBJ. DESC.=object descriptions; SIM./SENS.=simulator/sensors.
\end{tablenotes}
\end{threeparttable}
\end{table}

\noindent \textbf{Goal-Driven Navigation Datasets.}
Goal-driven navigation datasets have evolved to tackle increasingly sophisticated challenges, from basic spatial navigation to complex object interactions and multi-goal reasoning. Early efforts, such as the \textbf{Matterport3D} dataset \cite{Matterport3D}, provide 3D indoor environments but contain artifacts such as missing surfaces. The \textbf{Habitat-Matterport 3D (HM3D)} dataset \cite{ramakrishnan2021hm3d} mitigates these issues by minimizing artifacts such as missing surfaces in the \textbf{Matterport3D} dataset and offering high-fidelity reconstructions for PointGoal navigation, thereby reducing the perception sim-to-real gap. Extending this foundation, the \textbf{HM3D-OVON} dataset \cite{yokoyama2024hm3dovondatasetbenchmarkopenvocabulary} introduces open-vocabulary Object Goal Navigation, challenging agents to navigate to unseen object categories. Similarly, the \textbf{Instance Object Navigation (ION)} dataset \cite{10.1145/3474085.3475575} refines the focus to navigating to specific object instances based on fine-grained attribute descriptions such as color or material. The \textbf{Multi-Object Navigation (MultiON)} dataset \cite{gireesh2023sequenceagnosticmultiobjectnavigation} further increases task complexity by requiring agents to navigate an ordered sequence of objects, thereby testing their memory and planning capabilities. However, these datasets leverage static environments, without requiring agents to interact with objects during navigation. The real-world navigation process should be more dynamic. To address this limitation, the \textbf{iGibson} dataset \cite{8954627,9636667,li2021igibson20objectcentricsimulation} introduces Interactive Navigation, where agents must physically manipulate objects, such as pushing objects out of the path, to reach targets in cluttered indoor settings. Recently, with the advent of Large Language Models, the \textbf{DIVSCENE} dataset \cite{wang2024divscenebenchmarkinglvlmsobject} leverages the GPT-4-based \textbf{HOLODECK} system \cite{Yang_2024_CVPR} to generate large-scale, diverse environments with expert trajectories produced via breadth-first search (BFS), offering varied indoor scenes that enhance agent generalization across different environments.

\noindent \textbf{Task-Driven Navigation Datasets.}
Task-driven navigation datasets have advanced from foundational instruction-following benchmarks to interactive frameworks for enhanced real-world applicability. The \textbf{Room-to-Room (R2R)} dataset \cite{mattersim}, built on the image scans of indoor environments in the \textbf{Matterport3D} dataset, establishes Vision-and-Language Navigation (VLN) by requiring agents to follow natural language instructions in discrete environments. Subsequent works, such as the \textbf{VLN-CE} dataset \cite{krantz_vlnce_2020}, adapted these environments to continuous settings by using the 3D reconstructed environments from the scans used in \textbf{R2R}. Meanwhile, \textbf{VLN-CE-Isaac} \cite{cheng2024navila} further tailors the \textbf{VLN-CE} dataset for legged robots, incorporating LiDAR-based terrain adaptation. However, these datasets primarily utilize low-level, step-by-step instructions, requiring agents to execute commands without additional assistance. In contrast, the \textbf{Vision-based Navigation with Language-based Assistance (VNLA)} dataset \cite{nguyen2019vision} provides higher-level instructions (\eg \say{Find a towel in the kitchen}) and allows agents to request language-based guidance, shifting the focus to effective help-seeking strategies. Furthermore, datasets like \textbf{ALFRED} \cite{shridhar2020alfredbenchmarkinterpretinggrounded}, \textbf{DialFRED} \cite{Gao_2022}, and \textbf{TEACh} \cite{Padmakumar_Thomason_Shrivastava_Lange_Narayan-Chen_Gella_Piramuthu_Tur_Hakkani-Tur_2022} integrate navigation with object manipulation and dialogue. Specifically, \textbf{ALFRED} focuses on completing household tasks based on step-by-step instructions, \textbf{DialFRED} and \textbf{TEACh}, similar to \textbf{VNLA}, enhance \textbf{ALFRED} by allowing dialogue for clarification. Alongside these efforts, the \textbf{A-EQA} dataset \cite{OpenEQA2023} is proposed to address Embodied Question Answering (EQA), necessitating that agents explore environments to answer open-vocabulary questions. More recently, \textbf{OceanGym} \cite{oceangym2025} extends embodied navigation benchmarks to underwater AUV scenarios. It provides an approximately 800m\(\times\)800m marine environment with shallow and deep water settings, six-direction RGB and sonar observations, 85 perception task samples, and eight decision/navigation task domains, including locating underwater objects, searching for shipwrecks or aircraft debris, inspecting oil pipelines and wind turbines, and docking. Experiments show large gaps between MLLM-driven agents and human operators, particularly under low illumination and sonar-dependent conditions, highlighting the difficulty of underwater perception, memory, and long-horizon planning.

\subsection{Evaluation Metrics} \label{sec:nav_evaluation-metrics}

\begin{figure*}[ht]
    \centering
    \begin{minipage}[t]{0.48\textwidth}
        \centering
        \includegraphics[width=\linewidth]{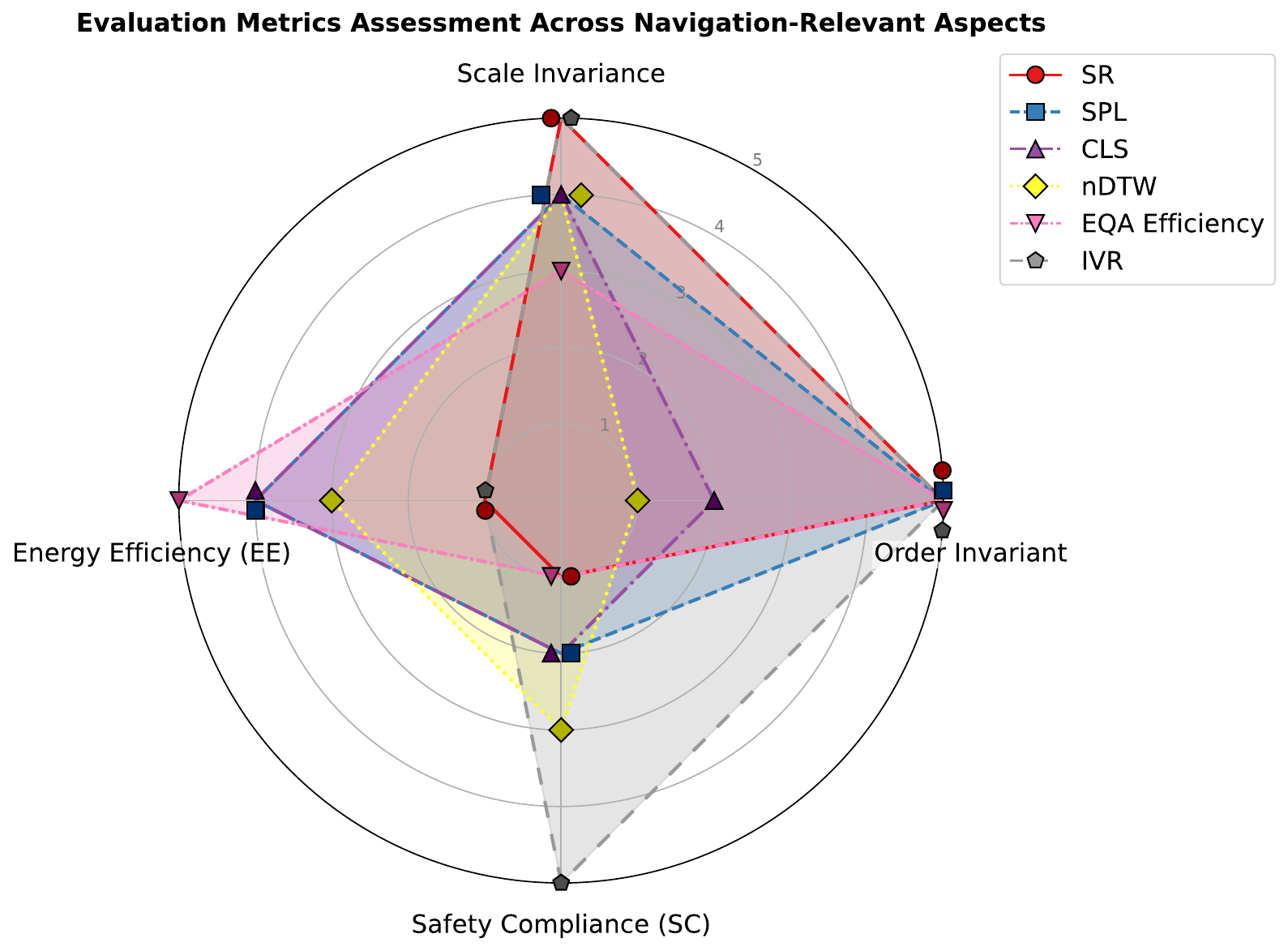}
        \vspace{-2mm}
        \centerline{\footnotesize (a) Metric-property coverage}
    \end{minipage}\hfill
    \begin{minipage}[t]{0.48\textwidth}
        \centering
        \includegraphics[width=\linewidth]{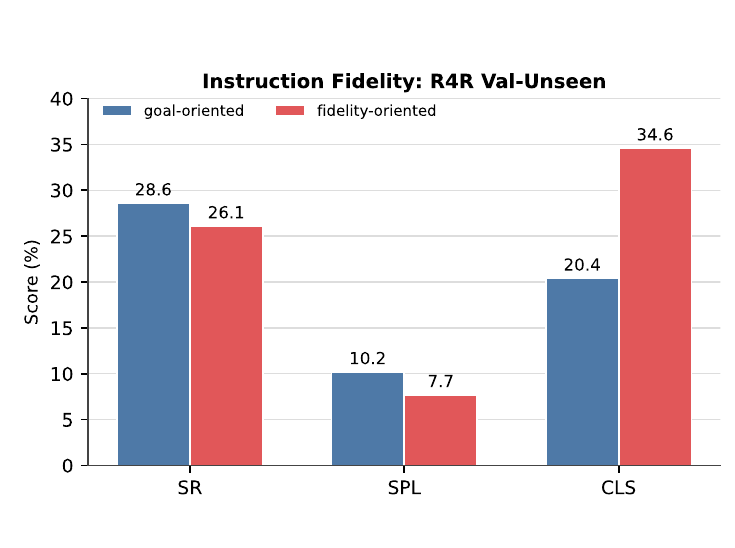}
        \vspace{-2mm}
        \centerline{\footnotesize (b) CLS stress test with SR and SPL on R4R val-unseen}
    \end{minipage}
    \caption{Navigation metrics. \textbf{(a)} Qualitative 1--5 coverage of scale/order invariance, safety, and energy efficiency. \textbf{(b)} R4R diagnostic: fidelity-oriented training improves CLS more than SR/SPL \cite{jain2019staypathinstructionfidelity}.}
    \vspace{-4mm}
    \Description{A radar chart and a bar chart. The left radar chart compares navigation metrics across scale invariance, order invariance, safety compliance, and energy efficiency. The right bar chart compares goal-oriented and fidelity-oriented agents on R4R validation-unseen using SR, SPL, and CLS.}
    \label{fig:nav_radar}
\end{figure*}

Each type of navigation task necessitates evaluation metrics tailored to its distinct objectives. Goal-driven navigation focuses on efficiently reaching predefined targets and is typically measured by quantitative indicators such as Success Rate (SR) and Path Length (PL). In contrast, task-driven navigation requires agents to either execute actions based on textual instructions or respond to questions after exploring the environment, thus demanding metrics that assess action-instruction alignment and answer accuracy. This section analyzes these metrics, exploring their properties across both task types. Additionally, we evaluate these metrics through four critical perspectives---Scale Invariance, Order Invariance, Safety Compliance, and Energy Efficiency---as illustrated in Figure \ref{fig:nav_radar}(a). These perspectives address key real-world considerations: Scale Invariance ensures metric robustness across diverse environment scales, from indoor settings to outdoor landscapes; Order Invariance allows agents to achieve the navigation goal via different paths, as multiple valid paths often exist in the real world; Safety Compliance assesses the ability of an agent to avoid collisions and hazardous zones; and Energy Efficiency evaluates resource optimization, often approximated via path length or navigation duration. Furthermore, figure \ref{fig:nav_radar}(b) uses a result from a navigation experiment reported in Jain et al. \cite{jain2019staypathinstructionfidelity}, where a fidelity-oriented reward substantially raises Coverage weighted by Length Score (CLS) while leaving SR or Success weighted by Path Length (SPL) in a much narrower range. This shows that goal completion and instruction fidelity can diverge. Moreover, Ilharco et al. make the same point from different angles: Normalized Dynamic Time Warping (nDTW) / Success weighted by normalized Dynamic Time
Warping (SDTW) correlates better with human path preferences than PL, Navigation Error (NE), CLS, SPL, and Success weighted by Edit Distance (SED) \cite{ilharco2019generalevaluationinstructionconditioned}.

\noindent \textbf{Metrics for Goal-Driven Navigation Tasks.}
In goal-driven navigation, where agents aim to reach targets, evaluation metrics prioritize both task completion and efficiency. The \textbf{Success Rate (SR)} measures the proportion of episodes in which the agent reaches the goal, offering a foundational indicator of performance. For object goal navigation, the \textbf{Instance-Localization Success Rate (ILSR)} \cite{10.1145/3474085.3475575} extends SR by requiring the agent not only to approach the target object but also to identify it among other visible instances correctly---for example, distinguishing a red chair from a blue one in a cluttered room---thereby assessing both localization and recognition capabilities. To incorporate path efficiency, the \textbf{Success weighted by Path Length (SPL)} \cite{anderson2018evaluationembodiednavigationagents}, defined as,
\begin{equation}
    \text{SPL} = \frac{1}{N} \sum_{i=1}^N S_i \frac{\ell_i}{\max(p_i, \ell_i)},
\end{equation}
where \(N\) is the number of episodes, \(S_i\) is the success indicator, \(\ell_i\) is the shortest path length, and \(p_i\) is the actual path length, balances success with a penalty for inefficiency.

\noindent \textbf{Metrics for Task-Driven Navigation Tasks.}
Task-driven navigation includes tasks like following textual instructions or answering questions after exploration. Tasks such as Vision-and-Language Navigation (VLN) may require agents to follow step-by-step instructions to navigate a specific path. Therefore, these tasks require metrics that compare the predicted path with the reference path. For example, the \textbf{Coverage weighted by Length Score (CLS)} \cite{jain2019staypathinstructionfidelity} assesses path alignment with a reference path. It is defined as \(\text{CLS}(P, R) = \text{PC}(P, R) \cdot \text{LS}(P, R)\), where Path Coverage \(\text{PC}(P, R)\) measures spatial coverage of the predicted path $P$ and reference path $R$ and Length Score (LS) measures how well the length of the predicted path matches the length of the reference path. Moreover, \textbf{Normalized Dynamic Time Warping (nDTW)} \cite{ilharco2019generalevaluationinstructionconditioned} extends this by considering both spatial alignment and the sequence of actions taken. It generates a similarity score ranging from 0 to 1, where a higher score reflects greater alignment with the reference path in terms of location and order. Both the \textbf{CLS} \cite{jain2019staypathinstructionfidelity} and the \textbf{nDTW} \cite{ilharco2019generalevaluationinstructionconditioned} ensure scale-invariance by normalizing the score by the reference path length. Furthermore, Song et al. \cite{song2024towards} break down the step-by-step instructions in VLN tasks into multiple subtasks and propose the \textbf{Independent Success Rate (ISR)} \cite{song2024towards} metric to evaluate completion of each subtask independently, enabling granular assessment of complex instructions like ``turn left, find the door''. In Embodied Question Answering (EQA), \textbf{EQA Efficiency} \cite{OpenEQA2023}, formulated as,

\begin{equation}
    E = \frac{1}{N} \sum_{i=1}^N \frac{(\sigma_i - 1)}{4} \times \frac{\ell_i}{\max(p_i, \ell_i)} \times 100\%.
\end{equation}

This metric integrates answer correctness (via an LLM-Match score \(\sigma_i\) from 1 to 5) with exploration efficiency, encouraging agents to gather accurate information through shorter paths. Finally, the \textbf{Instruction Violation Rate (IVR)} \cite{choi2024canvascommonsenseawarenavigationintuitive}, computed as \(\text{IVR} = \frac{\text{Number of episodes with violations}}{\text{Total number of episodes}}\), quantifies violations of commonsense constraints (\eg jaywalking), directly addressing compliance of the agents with the safety rules set by humans.

\subsection{Methods} \label{sec:nav_methods}

As illustrated in Figure \ref{fig:nav_overview}, navigation agents must overcome partial observability by building memory of historical observations and actions to guide their decisions. Memory can be categorized into Explicit and Implicit Memory.

\subsubsection{\textbf{Explicit Memory}}

Explicit memory methods construct structured representations of the environment for planning. They are divided into \textbf{Metric Map-Based} and \textbf{Graph-Based} approaches, which trade off between spatial precision and scalability.

\noindent \textbf{Metric Map-Based Methods.}
These methods discretize the environment into representations like grids, point clouds, or voxels to build a map, as shown in the 3D metric map in Figure \ref{fig:nav_overview}(a). Such maps are effective for fine-grained spatial reasoning and precise path-finding. For instance, many approaches use 2D occupancy grid maps to compute shortest paths \cite{ chaplot2020learningexploreusingactive}. To enhance environmental understanding, maps can also be augmented with semantic information, such as object labels or language-grounded features extracted by models such as CLIP \cite{chen2018learning, huang23vlmaps, radford2021learning}. While metric maps excel at capturing detailed spatial information for low-level control, their significant computational and memory requirements for maintenance and updates limit their scalability in large-scale environments. More recent methods, including VLFM \cite{yokoyama2024vlfm} and BeliefMapNav \cite{zhou2026beliefmapnav}, combine occupancy or frontier maps with language-grounded value or belief maps for zero-shot semantic navigation. These methods improve interpretability and open-vocabulary generalization but remain sensitive to dense memory costs, perceptual noise, and accumulated mapping errors.

\noindent \textbf{Graph-Based Methods.} 
To address the scalability limitations of metric maps, graph-based methods abstract the environment into a topological structure, where nodes represent key locations or landmarks and edges denote traversable paths, as seen in the hierarchical graph in Figure \ref{fig:nav_overview}(b). This abstraction is highly effective for efficient, long-range planning in large environments. Topological graphs are commonly used in tasks like Image-Goal Navigation, where nodes can represent panoramic views or image embeddings \cite{savinov2018semiparametrictopologicalmemorynavigation}. Additionally, knowledge graphs can model spatial and functional relationships between objects, enabling agents to infer object locations for semantic navigation tasks \cite{yang2018visualsemanticnavigationusing}. Path planning is then performed using classical algorithms like Dijkstra's or learned planners based on Graph Neural Networks (GNNs).

\subsubsection{\textbf{Implicit Memory}} 
Implicit memory methods rely on learned representations, pre-trained knowledge, or predictive models to infer actions from a history of observations. This approach comprises three categories: \textbf{Latent Representation-Based}, \textbf{Foundation Model-Based}, and \textbf{World Model-Based}.

\noindent \textbf{Latent Representation-Based Methods.}
These methods encode sequences of observations and actions into latent vectors, which serve as the agent's memory state, as depicted in Figure \ref{fig:nav_overview}(c). Architectures such as LSTMs and Transformers are used to process sequential inputs (\eg images, text) and maintain internal state, thereby integrating temporal information to mitigate partial observability \cite{zhu2020visiondialognavigationexploringcrossmodal, Hong_2021_CVPR}. While effective for short-horizon tasks, these methods compress historical information, potentially leading to the loss of fine-grained spatial details. This makes them prone to accumulating errors over long trajectories in complex environments.

\noindent \textbf{Foundation Model-Based Methods.}
Foundation model-based navigation leverages LLMs and VLMs primarily as semantic priors and high-level decision modules to interpret free-form goals into actionable subgoals (Figure \ref{fig:nav_overview}(d)). An early paradigm converts perception into textual descriptions, enabling an LLM to perform stepwise reasoning and action selection \cite{zhou2023navgpt,zheng2023learning}; newer approaches use VLM embeddings for open-vocabulary goal specification and scene understanding. For example, OpenFMNav \cite{kuang2024openfmnav} combines LLM-based parsing of user demands with VLM-based open-set perception to construct a semantic score map for exploration and exploitation, while OVExp \cite{wei2024ovexp} projects VLM features into top-down maps to support open-vocabulary, goal-conditioned exploration across unseen objects and goal modalities. To reduce partial-observability errors and hallucinations, recent works have coupled VLMs with structured memories (\eg semantic maps or retrieval history buffers) or adapted them to incorporate past experience via fine-tuning \cite{zhang2025mapnav}. More recently, several works have treated navigation itself as a primary pre-training objective and built navigation-specific foundation models. InternVLA-N1 \cite{internvla-n1} uses a dual-system design with waypoint planning and low-latency control, while NavA$^3$ \cite{zhang2025nava} pairs a global Reasoning VLM with a local point generation VLM for open-vocabulary, spatially grounded localization. Additionally, NavFoM \cite{zhang2025embodied} targets scalability across tasks and embodiments by unifying multi-view histories with viewpoint/time tokens and history selection. Meanwhile, NavRAG \cite{wang2025navrag} scales supervision by using retrieval-augmented LLM generation to produce user-style VLN instructions from hierarchical scene descriptions.

\noindent \textbf{World Model-Based Methods.}
World model-based methods learn a predictive dynamics model to forecast observations (or latent states) conditioned on candidate actions, enabling look-ahead planning, trajectory ranking, and synthetic data generation (Figure \ref{fig:nav_overview}(e)). For example, Navigation World Models \cite{bar2024navigationworldmodels} use controllable video generation to simulate egocentric rollouts for planning and scoring, while latent-dynamics approaches such as X-Mobility \cite{liu2024xmobilityendtoendgeneralizablenavigation} predict in a latent space and decouple world modeling from policy learning to improve generalization. Other efforts in autonomous driving, including the GAIA series \cite{gaia} and NVIDIA Cosmos \cite{nvidia2025cosmosworldfoundationmodel}, likewise emphasize controllable future generation and offline ``what-if'' evaluation of action predictions over observations.
\section{Manipulation} \label{sec:mani}

Successful sim-to-real transfer for manipulation requires agents to perceive task-relevant geometric and contact details, and simulators to model the physical dynamics that determine object motion, contact stability, and force interactions. Consider a robot learning to unscrew a bottle cap in a simulated environment: it must perceive and understand the shapes and poses of both the cap and its end effectors. Additionally, the simulator must simulate realistic physics interactions, including multi-point contacts, friction, and collision forces, to enable successful sim-to-real transfer of this task. As a result, current manipulation research focuses on advancing perception modeling and on developing simulators and physics engines. Notably, differentiable simulators have recently gained prominence. They provide gradients with respect to physical states, enabling training of policies with improved sim-to-real transferability.
Moreover, manipulation tasks vary in complexity. As tasks become more complex, agents require more advanced sensing modalities, perception methods, and hardware. Simpler tasks, such as planar grasping, can often be addressed using RGB cameras and basic grippers. However, dexterous in-hand manipulation requires tactile sensing, leveraging representations such as point clouds or voxels, and multi-fingered hands. This highlights that the requirements for perception, representation, simulators, and hardware vary with task difficulty.

\subsection{Manipulation tasks} \label{sec:mani_tasks}

\begin{figure*}[ht]
    \centering
    \includegraphics[width=\linewidth]{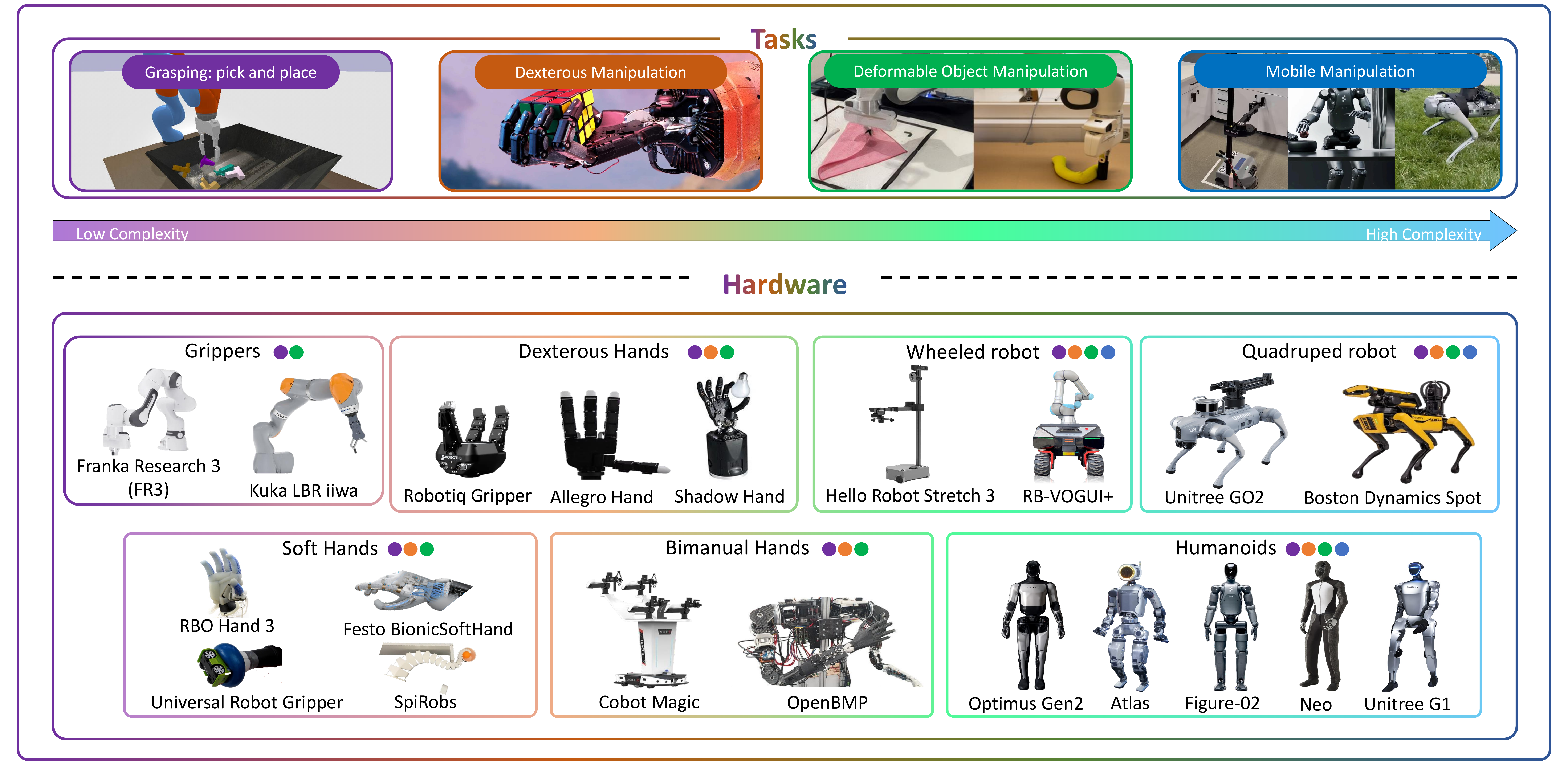}
    \caption{Manipulation tasks (top) and hardware (bottom), ordered by increasing complexity and DoFs.} 
    \vspace{-3mm}
    \Description{A two-part overview showing manipulation task categories and robot hardware ordered by increasing complexity and degrees of freedom.}
\label{fig:mani_overview}
\end{figure*}

To better understand challenges associated with manipulation, we first categorize key manipulation tasks and their associated hardware by complexity and required degrees of freedom (DoFs), as illustrated in Figure \ref{fig:mani_overview}.

\noindent \textbf{Grasping} is a fundamental task in robotics, often framed as a pick-and-place operation where a robot must hold an object and place it in a new position. Planar grasping, which involves three degrees of freedom (DoFs), is typically used for objects on flat surfaces. Meanwhile, full 3D grasping requires six DoFs---x, y, z, roll, pitch, and yaw---to handle objects in arbitrary poses, necessitating robotic arms with higher DoFs for effective coordination. Importantly, multi-fingered or dexterous hands can also be used for grasp acquisition, where the additional hand DoFs affect contact selection, internal-force regulation, and grasp stability/quality \cite{ zheng2025survey}. In this survey, we treat such cases as grasping when the primary objective is stable object acquisition or holding, and reserve ``dexterous manipulation'' for post-grasp reorientation or continuous in-hand object motion.

\noindent \textbf{Dexterous manipulation} requires controlling the pose and forces of an object through contact interactions. While often associated with in-hand manipulation using multi-fingered hands, it also encompasses extrinsic dexterity, in which simple grippers leverage environmental constraints to reorient objects. In the context of multi-fingered hands, this task employs techniques such as finger gaiting, rolling, or pivoting to control the orientation of the object in hand (\eg twisting a Rubik’s cube and spinning a pen). This task demands precise coordination to handle the complex contact dynamics involved. For objects with complex shapes, accurate simulation of frictional forces and multi-point contacts is essential---a capability provided by simulators such as MuJoCo \cite{mujoco}.

\noindent \textbf{Deformable object manipulation} involves handling soft materials, such as cloth or ropes, whose shape can change continuously when subjected to external forces. Unlike rigid objects, whose shape is fixed, deformable objects exhibit an effectively infinite number of degrees of freedom (their state space contains infinitely many possible configurations) because the relative distances between points in soft-body objects are not fixed, resulting in a highly dynamic and complex object state space. Tasks such as knot-tying or folding clothes require real-time monitoring of object geometric deformations and precise control that adapts to material properties, such as elasticity and friction.

\noindent \textbf{Fragile object manipulation} refers to grasping, lifting, transporting, or placing an object whose admissible contact loading is bounded above by a damage (crushing/breakage) threshold and bounded below by the minimum loading required to prevent slip \cite{s25175430}. Therefore, these tasks require precise force control and careful handling to prevent object damage. A common approach is to use soft robotic grippers made of materials such as rubber, silicone, or thermoplastic polyurethane (TPU). These grippers may employ actuators---such as pneumatic \cite{su2022pneumatic}, hydraulic \cite{10771693}, or tendon-driven \cite{WANG2025100646} systems---to control finger motion, ensuring uniform pressure distribution for safer handling. Precise force control is critical to avoiding damage, necessitating real-time feedback and adaptive control strategies tailored to varying levels of fragility (\eg eggshells and berries). Accurate simulation of object properties is essential for effective training of this task. Additionally, detecting fragility through visual cues (\eg geometry, texture) or tactile feedback is crucial for adjusting grip strength appropriately.

\noindent \textbf{Mobile manipulation} involves robot arms mounted on mobile platforms with navigation capabilities, such as wheeled bases, quadrupeds, or humanoids. In this setting, a robot must both navigate the environment and manipulate objects, such as navigating to a kitchen to pick up a cup.

\noindent \textbf{Open-world manipulation} tackles the \say{infinite variability problem} \cite{MITMani}. It requires robots to handle novel objects in unstructured and dynamic environments, such as picking up unseen items in cluttered spaces. The unpredictability of these settings requires robots to generalize from limited training data and adapt to new objects, materials, or conditions.

\noindent \textbf{Bimanual manipulation} employs dual-arm systems, such as ALOHA \cite{zhao2023learning}, to perform tasks that require coordination beyond the capabilities of a single arm, such as assembling Lego pieces.

\subsection{Physics Engines and Simulators}\label{sec:mani_sim}

\begin{figure}[t]
    \includegraphics[width=0.8\linewidth]{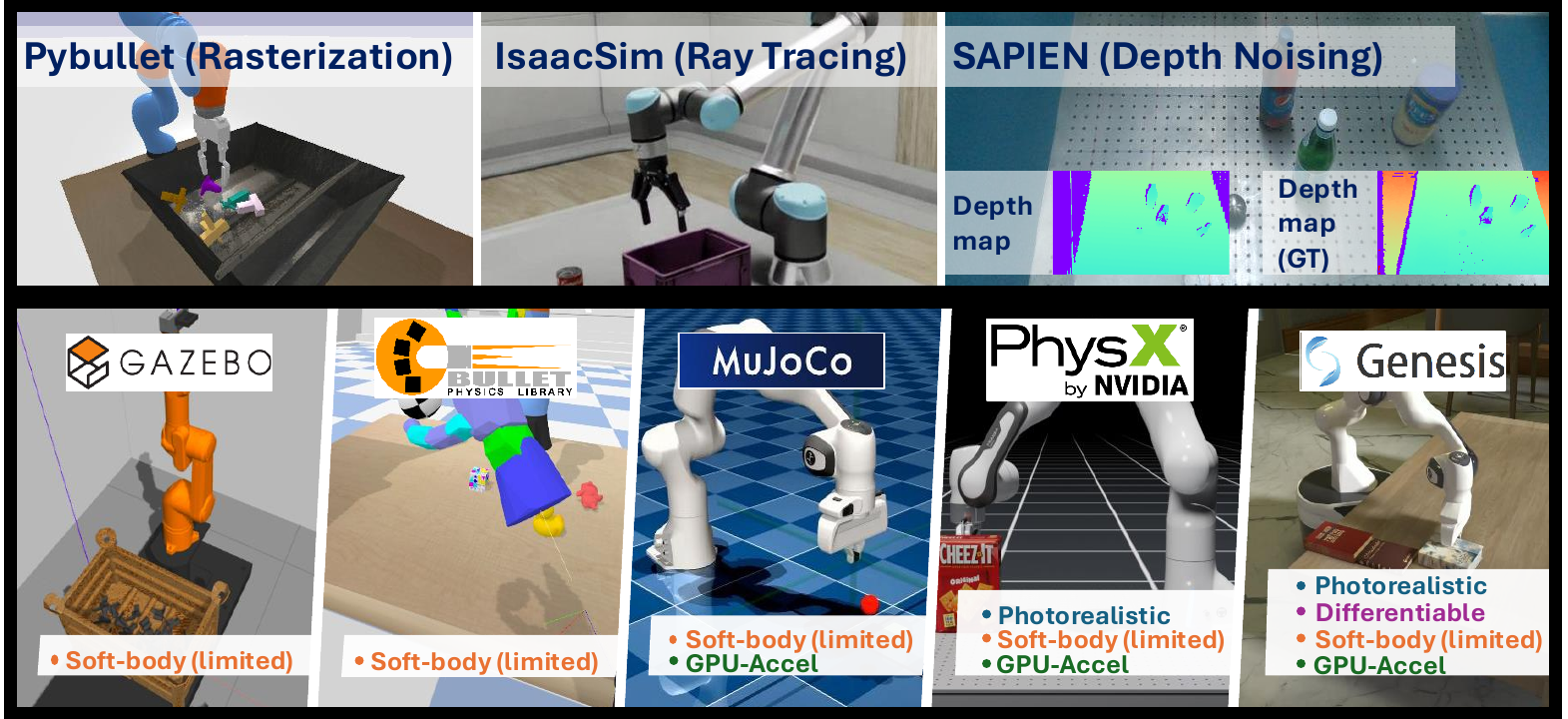}
    \centering
    \caption{\textbf{Upper}: Simulator rendering. PyBullet (left) uses rasterization, Isaac Sim (middle) employs photorealistic ray tracing, and SAPIEN (right) improves realism through depth noising. \textbf{Bottom}: Major physics engines and simulators, from Gazebo and early PhysX, MuJoCo, and PyBullet to newer platforms such as PhysX 5 and Genesis, highlighting advances in GPU acceleration, photorealistic rendering, and differentiable physics.}
    \vspace{-5mm}
\Description{A manipulation simulator comparison showing rendering examples and a timeline of major physics engines and simulators.}
\label{fig:mani_sim}
\end{figure}

Similar to navigation simulators, manipulation simulators must mitigate both perception and action-dynamics sim-to-real gaps for sim-to-real transfer. The action-dynamics gap is especially critical because manipulation involves more physical interaction between simulated objects and the workspace, requiring accurate modeling of contacts, material friction (\eg stone or ice), and collisions. Differentiable simulators (\eg Genesis/Genesis World \cite{genesis}) can facilitate gradient-based optimization of actions, controllers, or physical parameters, thereby improving sample efficiency and sim-to-real transfer when the modeled dynamics are sufficiently accurate. Recent efforts, such as PhysX-3D~\cite{caophysx}, improve realism by physically grounding assets (\eg scale, material, affordance, and kinematics). To reduce the perception gap, modern simulators adopt photorealistic rendering (\eg ray tracing) and sensor models such as depth noise. In this section, we compare classical engines (\eg MuJoCo \cite{mujoco}, Isaac Sim \cite{isaacsim}) with differentiable ones (\eg Dojo \cite{howelllecleach2022}, Genesis) and summarize key milestones and platform differences (Figure \ref{fig:mani_sim}, Table \ref{tab:mani_sim_decision} and \ref{tab:simulator_comparison}).

\noindent \textbf{Classical Physics Engines and Simulators.}
Classical simulators, which rely on traditional mechanics for simulating dynamics and contact modeling, have been foundational in robotic simulation for years. Below, we discuss several key classical simulators and their approaches to these physics and visual sim-to-real challenges.
\textbf{Gazebo} \cite{gazebo} integrates closely with the Robot Operating System (ROS) and supports multiple physics engines such as \textbf{Dynamic Animation and Robotics Toolkit (DART)}, \textbf{Open Dynamics Engine (ODE)}, and \textbf{Bullet} for rigid-body dynamics. These physics engines primarily excel in rigid-body simulation. Visually, it leverages the \textbf{Object-Oriented Graphics Rendering Engine (OGRE)}, which supports GPU-accelerated shading but lacks ray tracing or photorealistic capabilities. 
Meanwhile, \textbf{PyBullet} \cite{coumans2016pybullet} focuses on speed and efficiency. Built on the \textbf{Bullet} Physics engine, \textbf{PyBullet} provides GPU acceleration and continuous collision detection to enhance simulation speed. It uses a linear complementarity problem (LCP) contact model for physics simulation, which can be computationally intensive and may inaccurately approximate friction cones at contact points, impacting the action-dynamics sim-to-real gap. Rendering in \textbf{PyBullet} is limited to rasterization via \textbf{OpenGL}, without support for ray tracing or depth noise simulation, thus limiting its ability to narrow the perception sim-to-real gap. 
In contrast, \textbf{MuJoCo} \cite{mujoco} prioritizes precision in contact dynamics, making it useful for dexterous manipulation. It excels in multi-joint system simulation with accurate contact modeling, capturing robotic manipulator dynamics using generalized coordinates, and enabling stable, friction-rich interactions. However, its soft contact model can lead to interpenetration during impacts. Visually, it uses rasterization-based rendering via \textbf{OpenGL}, which lacks hardware-accelerated real-time ray tracing, restricting visual fidelity and widening the perception sim-to-real gap. However, it supports multi-threading for faster reinforcement learning. 
For photorealistic rendering, both \textbf{Isaac Sim} \cite{isaacsim} and \textbf{SAPIEN} \cite{Xiang_2020_SAPIEN,chang2015shapenet} stand out. They leverage GPU-accelerated rasterization and real-time ray tracing to create photorealistic environments with precise lighting and reflections, effectively reducing the perception sim-to-real gap. \textbf{SAPIEN} further supports built-in advanced depth noise simulation, which enhances visual fidelity by generating realistic depth maps with noise based on distance, object edges, and material properties, significantly improving sim-to-real transferability. They use the Nvidia \textbf{PhysX} engine for physics, providing robust simulation of rigid, soft, and fluid dynamics. 
Finally, \textbf{CoppeliaSim} \cite{coppeliasim} supports multiple physics engines (\textbf{MuJoCo}, \textbf{Bullet}, \textbf{ODE}, \textbf{Newton}, \textbf{Vortex}), enabling simulation of rigid, soft, and cloth dynamics. This adaptability helps tailor physics simulation to specific tasks, though its lack of GPU acceleration limits efficiency. Its rendering, primarily rasterization-based via \textbf{OpenGL} with partial ray-tracing support, falls short of the photorealism achieved by \textbf{Isaac Sim} and \textbf{SAPIEN}.

\begin{table*}[htbp]
\caption{
Decision-oriented comparison of manipulation simulators.
Ratings: High = suitable as a primary tool; Med. = requires tuning or extensions;
Low = not a typical first choice; Dep. = engine- or setup-dependent.
}
\label{tab:mani_sim_decision}
\centering
\footnotesize
\setlength{\tabcolsep}{4pt}
\renewcommand{\arraystretch}{1.05}   
\begin{threeparttable}
\begin{tabularx}{\textwidth}{@{}
L{1.8cm}
L{2.6cm}
Y
Y
Y
@{}}
\toprule
\textbf{Simulator}
& \textbf{Capabilities}
& \textbf{Suitable Tasks/Modeling}
& \textbf{Advantages}
& \textbf{Failure Modes} \\
\midrule

\textbf{MuJoCo}~\cite{mujoco}
& \makecell[tl]{Contact: High \\ Cloth/Soft: Med. \\ Vision: Low \\ Deploy: Med.}
& \textbullet~ Contact-rich control \newline \textbullet~ Dexterous manipulation \newline \textbullet~ Locomotion--manipulation \newline \textbullet~ System identification
& \textbullet~ Fast, stable dynamics \newline \textbullet~ Strong contact modeling \newline \textbullet~ Rich actuator models
& \textbullet~ Limited sensor realism \newline \textbullet~ interpenetration \newline \textbullet~ Cloth sim-to-real lacks validation \\
\addlinespace[0.3ex]

\textbf{Isaac Sim / Isaac Lab}~\cite{isaacsim}
& \makecell[tl]{Contact: High \\ Cloth/Soft: Med.--High \\ Vision: High \\ Deploy: High}
& \textbullet~ Photorealistic vision tasks \newline \textbullet~ Synthetic data generation \newline \textbullet~ Industrial manipulation \newline \textbullet~ mobile manipulation \newline \textbullet~ Large-scale RL (GPU)
& \textbullet~ PhysX \& RTX rendering \newline \textbullet~ Domain randomization \newline \textbullet~ Replicator data tools \newline \textbullet~ ROS 2 integration
& \textbullet~ High GPU/VRAM overhead \newline \textbullet~ Precision tasks sensitive to solver settings \newline \textbullet~ Steep learning curve \\
\addlinespace[0.3ex]

\textbf{SAPIEN / ManiSkill}~\cite{Xiang_2020_SAPIEN,mu2021maniskillgeneralizablemanipulationskill}
& \makecell[tl]{Contact: High \\ Cloth/Soft: Med. \\ Vision: High \\ Deploy: Med.}
& \textbullet~ Object-centric manipulation \newline \textbullet~ Articulated object \newline \textbullet~ Point-cloud/RGB-D policies \newline \textbullet~ Large-scale benchmarks
& \textbullet~ Efficient rendering \newline \textbullet~ GPU-parallelized visual data \newline \textbullet~ Strong articulated object support
& \textbullet~ Less deployment-focused than Gazebo \newline \textbullet~ Synthetic contact defaults need calibration \\
\addlinespace[0.3ex]

\textbf{PyBullet}~\cite{coumans2016pybullet}
& \makecell[tl]{Contact: Med. \\ Cloth/Soft: Low--Med. \\ Vision: Low \\ Deploy: Med.}
& \textbullet~ Lightweight prototyping \newline \textbullet~ Simple grasping \newline \textbullet~ Fast RL baselines \newline \textbullet~ Educational examples
& \textbullet~ Easy Python API \newline \textbullet~ Broad adoption \& low cost \newline \textbullet~ Useful collision queries
& \textbullet~ Contact sensitive to solvers \newline \textbullet~ Limited visual realism \newline \textbullet~ Unsuited for high-precision validation \\
\addlinespace[0.3ex]

\textbf{Gazebo}~\cite{gazebo}
& \makecell[tl]{Contact: Dep. \\ Cloth/Soft: Low \\ Vision: Med. \\ Deploy: High}
& \textbullet~ ROS/ROS 2 integration \newline \textbullet~ Mobile manipulation \newline \textbullet~ Pipeline validation \newline \textbullet~ SDF/URDF workflows \newline
& \textbullet~ Mature ROS integration \newline \textbullet~ SDF/URDF support \newline \textbullet~ Rich plugin ecosystem
& \textbullet~ Physics accuracy depends on engine plugin \newline \textbullet~ Unsuited for parallel RL \newline \textbullet~ Unsuited for contact-rich tasks \\
\addlinespace[0.3ex]

\textbf{CoppeliaSim} \cite{coppeliasim}
& \makecell[tl]{Contact: Dep. \\ Cloth/Soft: Low--Med. \\ Vision: Med. \\ Deploy: High}
& \textbullet~ Rapid prototyping \newline \textbullet~ Robot education \newline \textbullet~ Cross-engine comparisons
& \textbullet~ Flexible GUI \newline \textbullet~ Supports multiple physics engines
& \textbullet~ Results vary across engines \newline \textbullet~ Weak in large-scale RL \newline \textbullet~ Sim-to-real depends heavily on solver choice \\
\addlinespace[0.3ex]

\textbf{Genesis}~\cite{genesis}
& \makecell[tl]{Contact: Med.--High \\ Cloth/Soft: High \\ Vision: Med.--High \\ Deploy: Emerging}
& \textbullet~ Differentiable simulation \newline \textbullet~ Multi-physics experiments \newline \textbullet~ Soft/fluid/rigid interactions
& \textbullet~ Unified multi-physics design \newline \textbullet~ Differentiable support \newline \textbullet~ GPU-oriented architecture
& \textbullet~ Ecosystem is still developing \newline \textbullet~  Gradients through discontinuous contact can be brittle \\
\addlinespace[0.3ex]

\bottomrule
\end{tabularx}

\begin{tablenotes}[flushleft]
\footnotesize
\item \textbf{Capabilities:} Contact = contact-rich rigid manipulation; Cloth/Soft = cloth, rope, soft-body, compliant-gripper, or deformable-object tasks; Vision = photorealistic RGB-D/synthetic-data suitability; Deploy = convenience for ROS, hardware-in-the-loop, and real-robot pipelines.
\item ``Supports cloth/soft body'' should not be interpreted as ``validated for cloth folding sim-to-real transfer.'' Deformable manipulation additionally depends on material identification, self-contact handling, friction, damping, sensing, and controller-interface realism.
\end{tablenotes}
\end{threeparttable}
\end{table*}

Table~\ref{tab:mani_sim_decision} highlights that simulator selection is task-dependent.  For contact-rich articulated manipulation, MuJoCo is used for its efficient rigid-body dynamics, actuator modeling, and contact-rich control support.  For perception-heavy manipulation, Isaac Sim and SAPIEN/ManiSkill provide stronger rendering, RGB-D, point cloud, and synthetic data pipelines. For ROS-centered deployment, Gazebo and CoppeliaSim remain convenient, integrating naturally with robot middleware and hardware-oriented workflows. For cloth folding, rope manipulation, soft-body shaping, or compliant-gripper grasping, however, generic ``cloth'' or ``soft-body'' checkmarks are insufficient: the relevant question is whether the simulator models self-contact, bending/stretching, friction, damping, and material parameters at the fidelity needed by the target task. Specialized deformable simulators and benchmarks such as SoftGym, PlasticineLab, and IPC-based grasping simulators are better starting points for deformable-object research, while general-purpose platforms remain useful when perception, middleware, and full-robot deployment are the main concern.

\begin{table*}[htbp]
    \caption{\footnotesize Comparison of simulation platforms with sensing modalities. Abbreviations used: \textbf{Rendering} (Rast: Rasterization, RT: Ray Tracing, Pano.: panoramic/image-based rendering); \textbf{Dynamics} (R: Rigid, S: Soft, C: Cloth, F: Fluid). Data partially from \cite{Simulately}; names are hyperlinked. $^\star$No parallel gym support. $^\ddagger$Non-photorealistic only. $^\dagger$Supports Nvidia, AMD, Apple, \& TPU. $^{\blacklozenge}$Via 3rd-party (\eg ROS-X-Habitat). $^{\blacktriangle}$High-resolution visuotactile images are usually provided through external or sensor-specific libraries rather than generic native simulator sensors, e.g., TACTO \cite{wang2022tacto}, Taxim \cite{si2022taxim}, TacSL \cite{akinola2024tacsl}, TacEx \cite{tacex2024}, and Taccel \cite{taccel2025}. Native contact, touch, or force sensors should not be interpreted as calibrated GelSight/DIGIT-style tactile-image simulation. ROS: ROS Support, GPU: GPU accelerator support, OS: Open-Source.}
    \label{tab:simulator_comparison}
    \centering
    \scriptsize
    \setlength{\tabcolsep}{1.6pt}
    \renewcommand{\arraystretch}{0.96}
    \begin{threeparttable}
    \begin{tabularx}{\textwidth}{@{}
        L{0.17\textwidth}
        C{0.025\textwidth}
        C{0.16\textwidth}
        C{0.08\textwidth}
        C{0.065\textwidth}
        Y
        C{0.035\textwidth}
        C{0.035\textwidth}
        @{}}
    \toprule
    \textbf{Simulator} & \textbf{ROS} & \textbf{Physics Engine} & \textbf{Rendering} & \textbf{Dynamics} & \multicolumn{1}{c}{\textbf{Sensing modalities}} & \textbf{GPU} & \textbf{OS} \\
    \midrule
    \multicolumn{8}{@{}l}{\textit{Manipulation Simulators}} \\
    \midrule

    \textcolor{blue}{\href{https://sapien.ucsd.edu/}{SAPIEN}} \cite{Xiang_2020_SAPIEN}
    & \cmark
    & PhysX 5
    & Rast; RT$^\star$
    & R; S; F
    & RGB-D; Seg.; PC; Contact
    & \cmark
    & \cmark \\

    \textcolor{blue}{\href{https://developer.nvidia.com/isaac-sim}{Isaac Sim}} \cite{isaacsim}
    & \cmark
    & PhysX 5
    & Rast; RT
    & R; S; C; F
    & RGB-D; Seg.; LiDAR; Radar; IMU; Contact; F/T; Prox.; Tact.$^{\blacktriangle}$
    & \cmark
    &  \\

    \textcolor{blue}{\href{https://github.com/bulletphysics/bullet3}{PyBullet}} \cite{coumans2016pybullet}
    & \cmark
    & Bullet
    & Rast
    & R; S; C
    & RGB-D; Seg.; Ray; Contact; F/T; Tact.$^{\blacktriangle}$
    & 
    & \cmark \\

    \textcolor{blue}{\href{https://mujoco.org/}{MuJoCo}} \cite{mujoco}
    & \cmark
    & MuJoCo
    & Rast
    & R; S; C
    & RGB-D; Touch; F/T; IMU; Proprio.
    & \cmark$^\dagger$
    & \cmark \\

    \textcolor{blue}{\href{https://www.coppeliarobotics.com/}{CoppeliaSim}} \cite{coppeliasim}
    & \cmark
    & MuJoCo; Bullet; ODE; Newton; Vortex
    & Rast; RT$^\ddagger$
    & R; S; C
    & RGB-D; Prox.; Contact; F/T
    & 
    & \cmark \\

    \textcolor{blue}{\href{https://gazebosim.org/}{Gazebo}} \cite{gazebo}
    & \cmark
    & Bullet; ODE; DART; Simbody
    & Rast
    & R; S; C
    & RGB-D; Seg.; LiDAR; IMU; GPS; Mag.; Contact; F/T; DVL
    & 
    & \cmark \\

    \textcolor{blue}{\href{https://genesis-embodied-ai.github.io/}{Genesis}} \cite{genesis}
    & 
    & Genesis (DiffTaichi)
    & Rast; RT
    & R; S; C; F
    & RGB-D; Ray; IMU; Contact; F/T; Tact.
    & \cmark
    & \cmark \\

    \midrule
    \multicolumn{8}{@{}l}{\textit{Navigation Simulators}} \\
    \midrule

    \textcolor{blue}{\href{https://github.com/peteanderson80/Matterport3DSimulator}{Matterport3D}} \cite{Matterport3D}
    & 
    & --
    & Pano.
    & --
    & RGB-D; Pano.
    & \cmark
    & \cmark \\

    \textcolor{blue}{\href{https://aihabitat.org/}{Habitat}} \cite{habitat19iccv, szot2021habitat, puig2023habitat3}
    & \cmark$^{\blacklozenge}$
    & Bullet
    & Rast
    & R
    & RGB-D; Seg.; Audio; GPS/Compass
    & \cmark
    & \cmark \\

    \textcolor{blue}{\href{https://ai2thor.allenai.org/}{AI2-THOR}} \cite{kolve2022ai2thorinteractive3denvironment}
    & 
    & PhysX
    & Rast
    & R
    & RGB-D; Seg.; Meta.
    & \cmark
    & \cmark \\

    \textcolor{blue}{\href{https://svl.stanford.edu/igibson/}{iGibson}} \cite{xiazamirhe2018gibsonenv,li2021igibson20objectcentricsimulation}
    & \cmark
    & Bullet
    & Rast
    & R; S; C
    & RGB-D; Seg.; PC; Flow; LiDAR; Bump; Proprio.
    & \cmark
    & \cmark \\

    \textcolor{blue}{\href{https://carla.org/}{CARLA}} \cite{Dosovitskiy17}
    & \cmark
    & PhysX
    & Rast
    & R
    & RGB-D; Seg.; LiDAR; Radar; DVS; Flow; IMU; GNSS; Collision
    & \cmark
    & \cmark \\

    \textcolor{blue}{\href{https://microsoft.github.io/AirSim/}{AirSim}} \cite{airsim2017fsr}
    & \cmark
    & PhysX + FastPhysics
    & Rast
    & R
    & RGB-D; Seg.; Flow; LiDAR; IMU; GPS; Mag.; Baro.; Distance
    & \cmark
    & \cmark \\

    \textcolor{blue}{\href{http://www.threedworld.org/}{ThreeDWorld}} \cite{gan2021threedworldplatforminteractivemultimodal}
    & 
    & PhysX
    & Rast
    & R; S; C; F
    & RGB-D; Seg.; Flow; Audio; Collision; Force
    & \cmark
    & \cmark \\

    \textcolor{blue}{\href{https://github.com/patrykcieslak/stonefish}{Stonefish}} \cite{cieslak2019stonefish}
    & \cmark
    & Bullet (extended)
    & Rast
    & R; F
    & RGB; Sonar; DVL; IMU; GPS; Compass; Pressure; Acoustic
    & \cmark
    & \cmark \\

    \textcolor{blue}{\href{https://field-robotics-lab.github.io/dave.doc/}{Project DAVE}} \cite{zhang2022dave}
    & \cmark
    & Gazebo 11 (ODE; UUV Sim)
    & Rast
    & R; F
    & RGB; Sonar; LiDAR; DVL; USBL; IMU; Pressure
    & 
    & \cmark \\

    \textcolor{blue}{\href{https://robots.et.byu.edu/holoocean/}{HoloOcean}} \cite{potokar2022holoocean}
    & \cmark$^{\blacklozenge}$
    & Unreal (UE4/UE5; PhysX/Chaos)
    & Rast
    & R; F
    & RGB-D; Sonar; DVL; IMU; Pressure; GPS; Comm.
    & \cmark
    & \cmark \\

    \textcolor{blue}{\href{https://marusimulator.github.io/}{MARUS}} \cite{lonvcar2022marus}
    & \cmark
    & Unity3D (PhysX)
    & Rast
    & R; F
    & RGB; Depth; Sonar
    & \cmark
    & \cmark \\

    \textcolor{blue}{\href{https://github.com/open-airlab/UNav-Sim}{UNav-Sim}} \cite{amer2023unavsim}
    & \cmark
    & Unreal Engine 5 (Chaos) + AirSim
    & Rast
    & R; F
    & RGB-D; IMU; GPS; Mag.; Baro.
    & \cmark
    & \cmark \\

    \textcolor{blue}{\href{https://oceangpt.github.io/OceanGym/}{OceanGym}} \cite{oceangym2025}
    & \cmark$^{\blacklozenge}$
    & Unreal Engine 5.3 (HoloOcean-based)
    & Rast
    & R; F
    & RGB; Sonar
    & \cmark
    & \cmark \\

    \textcolor{blue}{\href{https://umfieldrobotics.github.io/OceanSim/}{OceanSim}} \cite{oceansim2025}
    & \cmark
    & PhysX (Isaac Sim)
    & Rast; RT
    & R
    & RGB; Sonar; DVL; IMU; Baro.
    & \cmark
    & \cmark \\

    \textcolor{blue}{\href{https://marine-gym.com/}{MarineGym}} \cite{marinegym2025}
    & \cmark
    & PhysX (Isaac Sim)
    & Rast; RT
    & R; F
    & RGB; Sonar; IMU; Depth
    & \cmark
    & \cmark \\

    \bottomrule
    \end{tabularx}
    \begin{tablenotes}[flushleft]
    \footnotesize
    \item \textbf{Sensor abbreviations:} RGB-D=RGB+Depth; Seg.=semantic/instance segmentation; PC=point cloud; Flow=optical/scene flow; Prox.=proximity; F/T=force--torque; Proprio.=proprioception; Mag.=magnetometer; Baro.=barometer; DVS=event camera; GNSS/GPS=global positioning; DVL=Doppler velocity log; USBL=ultra-short baseline acoustic positioning; Comm.=communication; Meta.=simulator object metadata; Tact. = tactile.
    \item \textbf{Sim-to-real interpretation:} visual modalities require camera/depth calibration and noise modeling; contact/F/T/tactile modalities require contact, friction, compliance, and controller-interface realism; underwater modalities additionally require turbidity, illumination, acoustic propagation, DVL dropout, pressure noise, and current modeling.
    \end{tablenotes}
    \end{threeparttable}
\end{table*}

\noindent \textbf{Differentiable Physics Engines and Simulators.}
Differentiable physics engines compute gradients of simulation states with respect to inputs---such as actions or object poses---thereby enabling backpropagation through physical interactions, including collisions and deformations. By modeling real-world physics with differentiable functions, these engines enable policies to be optimized for real-world performance directly within the simulation, aligning with fundamental physical principles. This approach enhances the adaptability and transferability of the trained policies to real-world applications.
\textbf{Dojo} \cite{howelllecleach2022} is a physics engine designed with optimization first principles. It improves contact modeling and provides gradient information for the kinematics of manipulation targets by formulating contact simulation as an optimization problem. By applying the implicit function theorem, \textbf{Dojo} provides smooth, differentiable gradients. 
Meanwhile, \textbf{DiffTaichi} \cite{hu2019difftaichi, hu2019taichi} is a programming language for differentiable simulators. It adopts the megakernel approach, merging multiple computation stages into a CUDA kernel to maximize GPU utilization and accelerate simulations.
\textbf{Genesis} \cite{genesis}, built on \textbf{DiffTaichi}, is an open-source simulator fully optimized for differentiable simulation. It supports gradient-based optimization of neural network controllers and the authors report 10 to 80 times faster simulation speeds than existing GPU-accelerated simulators while aiming to preserve physical fidelity. Additionally, Genesis includes a ray-tracing system for photorealistic rendering and a generative data engine that transforms natural language into multimodal data to autonomously generate task environments.

\subsection{Benchmark Datasets} \label{sec:mani_datasets}

\begin{figure*}[htbp] 
    \centering
    \includegraphics[width=\linewidth]{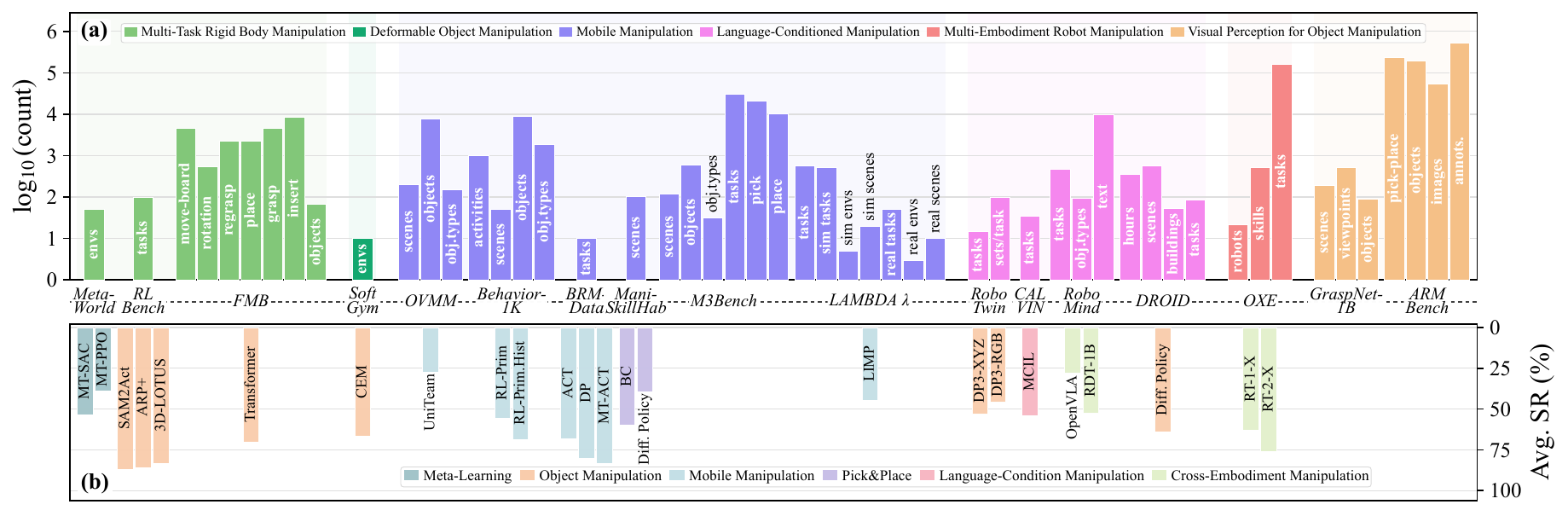}
    \caption{Manipulation benchmarks. \textbf{(a)} Dataset scale, measured by the benchmark's principal element (environments, tasks, scenes, or objects). \textbf{(b)} Representative average success rates, colored by evaluated task.}
    \Description{A two-panel bar chart comparing manipulation benchmark scale, dataset diversity, and representative method success rates.}
    \label{fig:manibenchmark}
\end{figure*}

This section provides an overview of benchmark datasets that are crucial for training agents that generalize across diverse tasks, environments, and robotic platforms. We categorize these benchmarks by the manipulation tasks they support, from rigid and deformable object manipulation to complex mobile and language-conditioned scenarios. Figure \ref{fig:manibenchmark} complements this overview by visualizing the scale, diversity, and method performance across these datasets, offering insights into data availability and current baseline capabilities.

\noindent \textbf{Rigid Body Manipulation Benchmarks.}
Several benchmarks are designed to explore the development of agents to handle rigid-body manipulation tasks. For instance, \textbf{Meta-World} \cite{yu2019meta} offers 50 distinct rigid-body manipulation environments, while \textbf{RLBench} \cite{james2019rlbench} contains 100 tasks. Both are designed to train agents with multiple basic skills and assess their generalization to new tasks. 

\noindent \textbf{Deformable Object Manipulation Benchmarks.}
Research on manipulating soft materials such as cloth and rope is supported by benchmarks such as \textbf{SoftGym} \cite{corl2020softgym} and \textbf{PlasticineLab} \cite{huang2021plasticinelab}. These platforms simulate the complex dynamics of deformable objects. Addressing their limitations in diversity, the \textbf{GRIP} dataset \cite{ma2025gripgeneralroboticincremental} introduces a more comprehensive benchmark featuring both soft and rigid grippers and a wide array of over 1,200 objects, simulated with high fidelity using an Incremental Potential Contact (IPC)-based simulator.

\noindent \textbf{Mobile Manipulation Benchmarks.} 
Mobile manipulation benchmarks evaluate an agent's ability to integrate navigation and manipulation for long-horizon tasks. \textbf{OVMM} \cite{yenamandra2024homerobotopenvocabularymobilemanipulation} challenges agents to move objects between locations in interactive 3D scenes. \textbf{BEHAVIOR-1K} \cite{li2022behavior} expands on this with 1,000 complex household activities in fully interactive environments. Similarly, \textbf{ManiSkill-Hab} \cite{shukla2024maniskillhab} focuses on long-horizon domestic chores, while \textbf{BRMData} \cite{zhang2024empowering} specifically targets bimanual mobile manipulation tasks.

\noindent \textbf{Language-Conditioned Manipulation Benchmarks.}
These benchmarks assess a robot's capacity to understand and execute natural-language instructions. \textbf{CALVIN} \cite{mees2022calvin} features long-horizon tasks requiring adherence to multi-step instructions. More recent large-scale, real-world datasets, such as \textbf{DROID} \cite{khazatsky2024droidlargescaleinthewildrobot} and \textbf{RoboMind} \cite{wu2024robomindbenchmarkmultiembodimentintelligence}, provide tens of thousands of real demonstration trajectories paired with linguistic annotations, fueling advances in imitation learning for instruction-following agents.

\noindent \textbf{Multi-Embodiment Robot Dataset.}
Recent benchmarks emphasize generalist policies transferable across hardware and environments. For example, \textbf{Open X-Embodiment} \cite{open_x_embodiment_rt_x_2023} provides data from 22 robot types for training X-robot policies, demonstrating 527 skills across 160,266 tasks. It is the largest open-source real-robot dataset, with over 1,000,000 task demonstration trajectories.

\noindent \textbf{Visual Perception Dataset.}
\textbf{GraspNet-1Billion} \cite{fang2023robust} is for grasping and perception tasks, such as 6D pose estimation and segmentation.
It includes 97,280 images, each annotated with precise 6D object poses and grasp points, covering 88 objects and offering over 1.1 billion grasp poses.

\subsection{Methods} \label{sec:mani_methods}
Robust manipulation in Embodied AI is shaped by the learning algorithm and the sensing modalities available to the robot. In practice, system design often begins by selecting task-relevant variables to observe and the bandwidth at which to observe them. For example, global object geometry and free space can often be captured by vision, whereas contact onset, slip, normal/shear force, and local compliance require force or tactile sensing. This sensing choice constrains the downstream representation, controller interface, simulator requirements, and the policy architecture itself. We therefore organize manipulation methods along two perspectives: (i) \textbf{Sensing Modalities}, which determine what information is observable and how quickly it can be fed back during execution, and (ii) \textbf{Policy Learning}, which determines how the resulting observations are mapped to actions.

\subsubsection{\textbf{Sensing Modalities}}
We categorize sensing modalities by the information they provide to the policy. Vision-based sensing captures global scene and object information, contact-based sensing captures local interaction states after contact, and vision--contact fusion combines both for contact-rich manipulation. 

\noindent \textbf{Vision-only (RGB / RGB-D / point cloud, etc.).}
Vision provides rich global context---including object identity, pose, geometry---and is therefore effective for geometry-dominant tasks such as reaching, pre-grasp planning, pick-and-place, and structured assembly when contact conditions are mild. In this regime, methods differ mainly in how they represent the scene for downstream control. Explicit geometric representations are widely used: scene-level discretizations (\eg voxel grids or point clouds) support spatial reasoning and region selection for downstream actions (\eg VoxPoser \cite{huang2023voxposercomposable3dvalue}, VoxAct-B \cite{liu2024voxactbvoxelbasedactingstabilizing}), while object-level abstractions such as 6D pose, grasp proposals, and affordance regions often suffice for grasp-centric pipelines. Recent works \cite{Park_2019,wen2024foundationposeunified6dpose,kerr2023lerflanguageembeddedradiance,rashid2023languageembeddedradiancefields,ji2024graspsplatsefficientmanipulation3d} also exploit language-conditioned querying of task-relevant parts and geometry-aware manipulation with neural 3D scene representations (\eg NeRF- \cite{kerr2023lerflanguageembeddedradiance,rashid2023languageembeddedradiancefields}/Gaussian-based \cite{ji2024graspsplatsefficientmanipulation3d} feature fields). Alternatively, implicit end-to-end visuomotor policies learn task-relevant representations directly inside the policy, avoiding an explicit intermediate state. However, vision-only manipulation remains fundamentally limited by the fact that images mainly observe external geometry rather than internal contact states. This limitation becomes most evident after contact is established: during grasp stabilization, vision may not reveal incipient slip, normal/shear force, or excessive squeezing; during sliding and surface following, small changes in friction, contact pressure, or contact mode can lead to jamming or loss of contact before any clear visual cue appears; and during deformable-object manipulation, self-occlusion and hidden contacts make it difficult to infer tension, wrinkles, or layer alignment. These failures are also temporal: many visual imitation policies predict action chunks for temporal consistency, but chunk execution is partially open-loop, so unexpected contact changes may not be corrected until the next policy inference. Temporal aggregation, as used in Action Chunking Transformer (ACT)-style policies \cite{fu2024mobilealohalearningbimanual}, can smooth overlapping action chunks, but it remains an inference-time post-processing mechanism and may trade reactivity for consistency. Thus, vision is highly effective for global scene understanding and pre-contact planning, but purely visual systems often struggle in contact-rich phases where success depends on local, high-frequency, and partially occluded interaction signals.

\noindent \textbf{Contact-only (force/torque, joint torque, tactile).}
Contact sensing is selected when the task-relevant state is primarily defined at the interaction interface rather than in the global visual scene. Force/torque and joint-torque signals reveal contact onset, load transfer, and force-regulation errors, and they naturally interface with impedance or hybrid force--position control. Tactile sensing further provides spatial contact structure, ranging from taxel arrays to optical tactile imagers (\eg GelSight \cite{yuan2017gelsight}, DIGIT \cite{lambeta2020digit}). Importantly, tactile representations are not limited to local geometry: by observing deformation and motion fields at the contact interface, vision-based tactile sensors can infer normal/shear/torsional loading and detect incipient slip \cite{yuan2015measurement_shear_slip_gelsight,yuan2017gelsight}, enabling closed-loop regulation of end-effector commands (\eg modulating grip force or gripper closure to prevent slip or avoid damaging fragile objects) \cite{li2024feelanyforce}. Learning-based methods can then map short tactile or force histories to corrective actions \cite{dong2021tactile}, supporting occlusion-robust contact-state estimation and deformable or dexterous manipulation~\cite{ma2021extrinsic_contact_sensing,yang2023tacgnn}. However, compared with vision, contact data are harder to scale: it requires physical interaction, careful calibration, and often additional instrumentation for force supervision, while tactile skins may exhibit wear and response drift \cite{navarroguerrero2023visuohaptic,bhirangi2021reskin}. These constraints motivate tactile simulation and synthetic data generation toolkits (\eg TACTO \cite{wang2022tacto}, Taxim \cite{si2022taxim}, TacSL \cite{akinola2024tacsl}). Contact-only sensing is also local and reactive: it provides limited pre-contact information for search and alignment, is sensitive to mounting and calibration drift, and can remain ambiguous under simultaneous multi-contact.

\noindent \textbf{Vision--contact fusion (Visuo-Tactile/Visual-Tactile).}
Visuo-tactile manipulation combines vision and tactile sensing to generate actions. This combination is particularly beneficial when vision guides objects approaching and coarse alignment, while contact stabilize execution under occlusion and during force-sensitive adjustments, including slip-aware grasp stabilization and insertion \cite{yuan2015measurement_shear_slip_gelsight,li2024feelanyforce}. Fusion can be performed at different levels. At the representation level, visual and tactile streams may be encoded jointly through cross-attention, or multimodal tokens; for example, Vision-Tactile-Language-Action Model (VTLA) \cite{zhang2025vtla} integrates visual and tactile images, and language instructions into a vision--tactile--language--action model for peg-in-hole insertion, using temporally enhanced tactile tokens and preference learning to improve continuous action prediction. At the control level, modalities may instead be assigned to different timescales: Reactive Diffusion Policy (RDP) \cite{xue2025reactive} uses a slow visual diffusion policy to generate coarse action chunks and a fast tactile/force pathway to correct actions within the chunk at high frequency, thereby preserving the trajectory-modeling benefits of action chunking while restoring closed-loop reactivity for contact-rich tasks. These examples suggest that multimodal fusion is not merely concatenating sensors; it also determines which modality controls which phase and bandwidth of the behavior. Beyond task-specific systems, paired visuo-tactile datasets and cross-modal objectives enable transferable multimodal representations that can be fine-tuned for downstream control (``Touch and Go'' \cite{yang2022touch}, cross-modal touch representation learning~\cite{zambelli2021learning}), and vision can provide supervision or rewards while tactile remains a robust execution-time observation channel (``See to Touch'' \cite{guzey2024see}).

\subsubsection{\textbf{Policy Learning}}

Recent advancements in manipulation focus on learning-based policies $\pi_\theta(\mathbf{a_{t}}|\mathbf{o_t})$. These methods have evolved significantly, from foundational learning paradigms to sophisticated, large-scale models. Yet, the following families are not mutually exclusive in practice (\eg initializing with imitation learning and improving with reinforcement learning, or combining Vision-Language-Action (VLA) models with diffusion-based action generation). Table \ref{tab:mani_equations} summarizes representative policy-learning families, compact formulations, and computational resource profiles.

\begin{table*}[htbp]
\caption{
Overview of policy learning methods for manipulation with representative computational complexity.
Parameter counts are order-of-magnitude examples; the dominant bottleneck may instead be rollout, planning, denoising, or GPU memory.
}
\label{tab:mani_equations}
\centering
\scriptsize
\setlength{\tabcolsep}{3pt}
\renewcommand{\arraystretch}{1.06}

\begin{threeparttable}
\begin{tabularx}{\linewidth}{@{}
L{0.055\linewidth}
>{\hsize=0.95\hsize\RaggedRight\arraybackslash}X
>{\hsize=1.05\hsize\RaggedRight\arraybackslash}X
@{}}
\toprule
\textbf{Family} 
& \textbf{Method, compact formulation, and references}
& \textbf{Resource profile} \\
\midrule

\multicolumn{3}{@{}l}{\textbf{Classical Policy Learning}}\\
\midrule

RL
& \meth
{Model-Free RL}
{\cite{openaidex,schulman2017proximalpolicyoptimizationalgorithms}}
{Learns a value function or policy directly from reward-driven interaction.}
{Q_\theta(\mathbf{s},\mathbf{a})\ \mathrm{or}\ \pi_\theta(\mathbf{a}\mid\mathbf{s})}
& \textbf{Parameter scale:} \(\sim 10^5\)--\(10^7\) for many task policies.

\textbf{Training bottleneck:} Dominated by rollout/simulation cost; often requires many environment interactions.

\textbf{Inference/deployment bottleneck:} Usually one policy forward pass; sim-to-real adaptation can add substantial cost.
\\
\addlinespace[0.6ex]

RL
& \meth
{Model-Based RL}
{\cite{nagabandi2019deepdynamicsmodelslearning,janner2021trustmodelmodelbasedpolicy,hafner2020dreamcontrollearningbehaviors,chua2018deepreinforcementlearninghandful,NEURIPS2018_2de5d166}}
{Learns dynamics, then plans or improves a policy.}
{\hat{p}_\phi(\mathbf{s}_{t+1}\mid\mathbf{s}_t,\mathbf{a}_t)}
& \textbf{Parameter scale:} \(\sim 10^5\)--\(10^7\) for many learned dynamics models.

\textbf{Training bottleneck:} More sample-efficient than model-free RL in many settings, but requires model learning.

\textbf{Inference/deployment bottleneck:} Online MPC, trajectory optimization, or model rollouts can dominate deployment cost.
\\

\midrule

IL
& \meth
{Behavior Cloning}
{\cite{wong2021errorawareimitationlearningteleoperation,torabi2018behavioralcloningobservation,florence2021implicitbehavioralcloning}}
{Supervised learning from observation--action demonstrations.}
{\theta^*=\arg\min_\theta\sum_i \ell(\pi_\theta(\mathbf{o}_i),\mathbf{a}_i)}
& \textbf{Parameter scale:} \(\sim 10^6\)--\(10^8\), depending on the visual encoder.

\textbf{Training bottleneck:} Collecting demonstrations.

\textbf{Inference/deployment bottleneck:} One forward pass; usually practical for limited compute.
\\
\addlinespace[0.6ex]

IL
& \meth
{Action Chunking / ACT}
{\cite{zhao2023learning,fu2024mobilealohalearningbimanual,cheng2024opentelevisionteleoperationimmersiveactive}}
{Predicts a short action horizon instead of a single action.}
{\pi_\theta(\mathbf{a}_{t:t+H}\mid\mathbf{o}_{\leq t})}
& \textbf{Parameter scale:} \(\sim 10^7\)--\(10^8\) for typical transformer/CVAE policies.

\textbf{Training bottleneck:} Requires demonstrations but avoids massive RL rollout.

\textbf{Inference/deployment bottleneck:} One forward pass predicts multiple actions, reducing effective control frequency.
\\
\addlinespace[0.6ex]

IL
& \meth
{Human Video Retargeting}
{\cite{nair2022r3muniversalvisualrepresentation,yang2024acecrossplatformvisualexoskeletonslowcost,he2024omnih2ouniversaldexteroushumantohumanoid,fu2024humanplushumanoidshadowingimitation,li2024okamiteachinghumanoidrobots}}
{Transfers human motion or pose estimates to robot actions.}
{\mathbf{a}=f_{\mathrm{retarget}}\!\left(f_{\mathrm{pose}}(\mathbf{o})\right)}
& \textbf{Parameter scale:} Backbone models may range from \(\sim 10^7\) to \(10^9+\).

\textbf{Training bottleneck:} Pose estimation, calibration, and embodiment retargeting.

\textbf{Inference/deployment bottleneck:} Pose inference plus retargeting optimization; embodiment mismatch can dominate.
\\
\addlinespace[0.6ex]

IL
& \meth
{Diffusion Policy}
{\cite{chi2024diffusionpolicyvisuomotorpolicy,liu2024rdt1bdiffusionfoundationmodel,ze20243ddiffusionpolicygeneralizable,ze2024generalizablehumanoidmanipulationimproved,huang20253dvitaclearningfinegrainedmanipulation,ren2024learninggeneralizable3dmanipulation}}
{Models action distributions through iterative denoising.}
{p_\theta(\mathbf{A}_t\mid\mathbf{O}_t)}
& \textbf{Parameter scale:} \(\sim 10^7\)--\(10^8\) for task policies; foundation variants can be \(10^9+\).

\textbf{Training bottleneck:} Denoising objectives.

\textbf{Inference/deployment bottleneck:} Inference cost scales with the number of denoising steps \(K\).
\\

\midrule

\multicolumn{3}{@{}l}{\textbf{Language-Conditioned Policies}}\\
\midrule

VLMs
& \meth
{Vision-Language Models}
{\cite{yang2024embodiedmultimodalagenttrained,driess2023palmeembodiedmultimodallanguage,Liu_2024,zhaxizhuoma2024alignbotaligningvlmpoweredcustomized}}
{Map visual observations and language instructions to symbolic or textual actions.}
{\pi(l_{\mathrm{act}}\mid\mathbf{o},l_{\mathrm{ins}})}
& \textbf{Parameter scale:} \(\sim 10^9\)--\(10^{11}+\) for large VLM/LLM backbones.

\textbf{Training bottleneck:} Full training or fine-tuning is expensive; frozen/API use is more accessible.

\textbf{Inference/deployment bottleneck:} Usually used as high-level planners and require a separate low-level controller.
\\
\addlinespace[0.6ex]

VLAs
& \meth
{Vision-Language-Action Models}
{\cite{rt12022arxiv,brohan2023rt2visionlanguageactionmodelstransfer,pan2024visionlanguageactionmodeldiffusionpolicy,wen2024diffusionvlascalingrobotfoundation,black2024pi0visionlanguageactionflowmodel,kim2024openvlaopensourcevisionlanguageactionmodel,nvidia2025gr00tn1openfoundation}}
{Predict robot actions directly from vision and language.}
{\pi_\theta(\mathbf{a}\mid\mathbf{o},l_{\mathrm{ins}})}
& \textbf{Parameter scale:} \(\sim 10^7\)--\(10^{10}+\), from compact transformers to foundation VLAs.

\textbf{Training bottleneck:} Training from scratch is expensive; LoRA, adapters, or released checkpoints are preferable.

\textbf{Inference/deployment bottleneck:} Deployment may need edge GPU or server-side inference; quantization can reduce memory.
\\

\bottomrule
\end{tabularx}

\begin{tablenotes}[flushleft]
\footnotesize
\item \textbf{Interpretation.}
Parameter count alone does not determine computational accessibility. For RL, rollout cost often dominates; for model-based RL, online planning can dominate; for diffusion policies, denoising steps dominate inference latency; and for VLMs/VLAs, GPU memory and fine-tuning cost are often limiting factors.
\item \textbf{Low-resource takeaway.}
BC and ACT-style imitation learning are typically the most accessible when demonstrations are available. Diffusion policies and VLAs can be made more practical through smaller backbones, fewer denoising steps, quantization, LoRA/adapters, or released checkpoints.
\end{tablenotes}

\end{threeparttable}
\end{table*}

\noindent \textbf{Reinforcement Learning (RL) and Imitation Learning (IL).} 
RL optimizes policies through trial and error, using either model-free approaches such as PPO \cite{schulman2017proximalpolicyoptimizationalgorithms} or model-based methods that learn environment dynamics \cite{nagabandi2019deepdynamicsmodelslearning}. However, the extensive interaction requirements of RL make it inefficient for complex real-world manipulation. Imitation Learning (IL) offers a more data-efficient alternative by learning from expert demonstrations collected via teleoperation or human videos \cite{zhao2023learning, nair2022r3muniversalvisualrepresentation}. A primary IL method, Behavior Cloning (BC), learns a direct mapping from observations to actions but can suffer from compounding errors. Action Chunking Transformer (ACT) \cite{zhao2023learning} mitigates this by predicting a horizon of future actions rather than a single action, improving temporal consistency for fine-grained bimanual manipulation. ACT also uses temporal aggregation/ensembling at inference time, which averages overlapping predictions for the same timestep to smooth execution. This aggregation is best understood as a post-processing mechanism. It improves continuity but does not, by itself, provide high-bandwidth feedback during chunk execution. Consequently, later work revisits action chunking under two complementary constraints: contact-rich tasks require reactive feedback within chunks, as in Reactive Diffusion Policy (RDP) \cite{xue2025reactive}, while large VLA models require latency-aware real-time execution of chunks, as in Real-Time Chunking (RTC) for flow/diffusion action policies \cite{black2025realtimechunking}. 

\noindent \textbf{Diffusion Policies (DPs).} As a powerful extension of IL, Diffusion Policies (DPs) \cite{chi2024diffusionpolicyvisuomotorpolicy} have gained prominence. They use denoising diffusion models to generate actions, offering stable training and the ability to model complex, multi-modal action distributions. This makes them highly effective for fine-grained manipulation tasks. Recent work has advanced DPs by conditioning them on 3D point cloud representations, which enhances generalization in complex scenes \cite{ze20243ddiffusionpolicygeneralizable}.

\noindent \textbf{Vision-Language Models (VLMs).}
VLM-based methods typically operate as decision-making or planning modules that translate visual observations and language instructions into subgoals, symbolic plans, or action descriptions, which are then executed by separate controllers~\cite{driess2023palmeembodiedmultimodallanguage, Liu_2024, zhaxizhuoma2024alignbotaligningvlmpoweredcustomized}. This decomposition can improve interpretability and modularity, but introduces grounding and interface challenges: the downstream controller must reliably execute the abstract plan under contact uncertainty and partial observability.

\noindent \textbf{Vision-Language-Action (VLA) models.}
VLAs directly output robot actions conditioned on observations and language instructions, thereby jointly learning representations and control. Early exemplars include RT-1~\cite{rt12022arxiv} and RT-2~\cite{brohan2023rt2visionlanguageactionmodelstransfer}, while large-scale multi-robot training (\eg RT-X~\cite{open_x_embodiment_rt_x_2023}) and open implementations such as OpenVLA~\cite{kim2024openvlaopensourcevisionlanguageactionmodel} have improved accessibility and cross-embodiment generalization. Moreover, systems increasingly combine architectural elements (\eg diffusion-based continuous action generation or dual-system designs that separate reasoning from low-latency control)~\cite{pan2024visionlanguageactionmodeldiffusionpolicy,nvidia2025gr00tn1openfoundation,black2024pi0visionlanguageactionflowmodel}. Recently, a growing body of work revisits RL as a practical post-training mechanism for VLAs to address long-tail failures, mixed-quality data, and real-world distribution shift. Recent directions include: (i) \textbf{policy-centric RL fine-tuning} of VLAs via online or hybrid RL schedules (\eg alternating RL and supervised updates) \cite{11127299}; (ii) \textbf{offline RL / value learning} objectives that extract robust behavior from heterogeneous demonstrations \cite{zhang2025reinbot}; (iii) \textbf{learned reward/critic models} based on VLM/VLA backbones to reduce reward engineering and provide dense progress signals \cite{lee2026roborewardgeneralpurposevisionlanguagereward}; and (iv) \textbf{reinforcement of reasoning and planning} components to improve long-horizon reliability \cite{huangthinkact}. These advances suggest that, while large-scale IL remains the dominant route to obtain a strong initial VLA policy, RL-based post-training is becoming important. Beyond vision and language alone, recent VLA systems also incorporate contact sensing. For example, Vision-Tactile-Language-Action (VTLA)~\cite{zhang2025vtla} conditions continuous action prediction on visual observations, tactile images, and language instructions for insertion manipulation, using tactile tokens and preference learning to generate contact-sensitive actions. 
\section{Discussion} \label{sec:discuss}

\subsection{Unification of Navigation and Manipulation}

The separation between navigation and manipulation is useful for organizing algorithms. However, many robotic tasks require an agent to possess both capabilities simultaneously. In household rearrangement, open-vocabulary mobile manipulation, and long-horizon activity benchmarks, an agent must both search through partially observed space, identify objects and receptacles, move its body to feasible interaction poses, grasp or push objects, transport them, and verify the final state \cite{szot2021habitat,yenamandra2024homerobotopenvocabularymobilemanipulation,li2022behavior,shukla2024maniskillhab}. In this view, navigation changes what can be perceived and reached, while manipulation changes the environment that later navigation and planning must reason over.

A first line of work treats unification as hierarchical system integration. These systems typically maintain an explicit spatial or semantic state, use a language model, vision-language model, symbolic planner, or value-map representation to decompose an instruction into subgoals, and then call specialized skills for exploration, navigation, grasping, placing, or articulated-object interaction \cite{ahn2022can,huang2023voxposercomposable3dvalue,zhou2023navgpt,zheng2023learning}. This design is practical because each module can use mature tools. For example, such systems can combine path planning for navigation, perception and grasp synthesis for manipulation, and low-level controllers for execution. However, the main limitation is the interface between modules. A high-level planner may produce a semantically correct subgoal that is unreachable, dynamically infeasible, or unsafe under the current contact and embodiment constraints. Conversely, a low-level skill may succeed locally but leave the scene in a state that violates the assumptions of the next skill \cite{szot2021habitat,yenamandra2024homerobotopenvocabularymobilemanipulation}.

A second line of work tries to reduce hand-designed interfaces by learning more shared action and representation spaces. Large robot datasets and VLA models, such as RT-1 \cite{rt12022arxiv}/RT-2 \cite{brohan2023rt2visionlanguageactionmodelstransfer}, RT-X \cite{open_x_embodiment_rt_x_2023}, OpenVLA \cite{kim2024openvlaopensourcevisionlanguageactionmodel}, and \(\pi_0\) \cite{black2024pi0visionlanguageactionflowmodel}, show that vision-language pretraining and multi-robot demonstration data can improve generalization for language-conditioned robot control. More recent latent-action approaches make the unification problem more explicit. For example, UniVLA \cite{bu2025univla} learns task-centric latent actions from videos and then decodes them into embodiment-specific controls, allowing manipulation, navigation, and egocentric human videos to contribute to a common policy representation. Cross-domain embodied VLMs, such as MiMo-Embodied \cite{hao2025mimoembodied}, similarly suggest that spatial reasoning, affordance prediction, and task planning can share a common multimodal backbone across domains, although these models remain closer to high-level reasoning modules than to complete closed-loop controllers. The important trend is therefore not only about scaling model size, but also about separating task intent from embodiment-specific execution. For example, language and vision specify what should change, latent actions or action tokens represent how progress should be made, and robot-specific controllers realize the motion under physical constraints.

Humanoid and legged mobile manipulation make these constraints clear. Whole-body tasks require the robot to coordinate base motion, balance, arm motion, hand contact, body posture, and sometimes foot-ground interaction. Recent systems such as WholeBodyVLA \cite{jiang2025wholebodyvla}, \(\Psi_0\) \cite{wei2026psi0}, HuMI \cite{nai2026humi}, and unified legged-manipulator policies \cite{hou2025efficientunified} therefore combine high-level visual-language or latent-action representations with embodiment-specific data collection, retargeting, RL controllers, or whole-body control interfaces. From the simulator-property perspective of this survey, this trend raises more fidelity requirements. For example, a simulator for unified agents must model not only traversability and visual semantics, but also object contacts, articulated objects, base-arm coupling, foot contacts, actuation limits, controller timing, sensor latency, and failure recovery.

\subsection{Error Analysis: Common Failure Cases in Simulator-Based Navigation and Manipulation}
\label{sec:error_analysis}

\begin{figure*}[htbp]
\centering
\includegraphics[width=\textwidth]{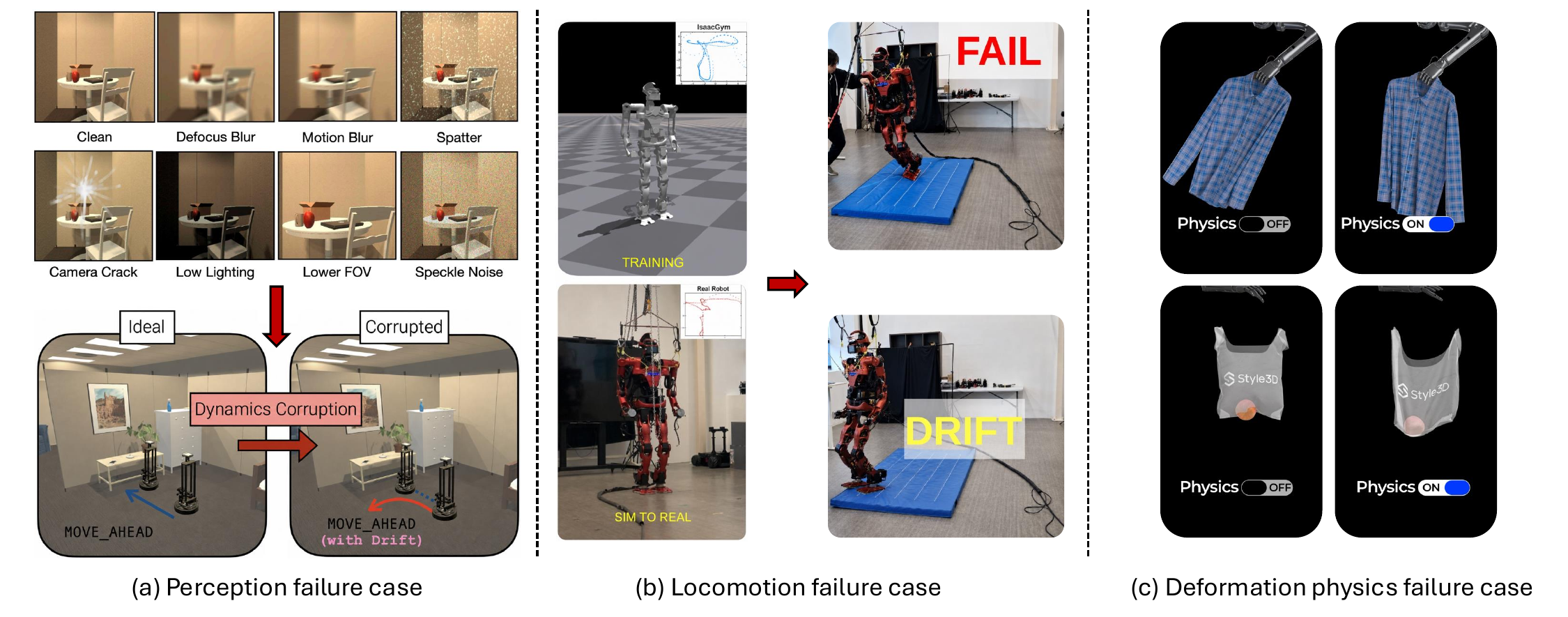}
\caption{Simulator-induced failures: \textbf{(a)} corrupted sensing or transitions \cite{chattopadhyay2021robustnav}; \textbf{(b)} locomotion mismatch in actuation, timing, balance, contact, terrain, or whole-body dynamics \cite{cha2025sim}; and \textbf{(c)} unrealistic deformable-object drape, stretch, friction, or self-contact (adapted from SynReal/Style3D \cite{synreal2026,style3d2026deformable}).}
\Description{A three-panel comparison of simulator-induced failure modes. Panel (a) shows visual observation corruptions, including blur, spatter, camera damage, low lighting, restricted field of view, and noise, together with an example of corrupted transition dynamics. Panel (b) contrasts simulated humanoid locomotion training with real-robot outcomes labeled failure and drift. Panel (c) compares deformable-object simulations in which alternative physics settings produce different cloth and object configurations.}
\label{fig:error_schematics}
\vspace{-1mm}
\end{figure*}

Simulation scales training and evaluation, but deployment still fails when policies depend on mismatched properties. Figure \ref{fig:error_schematics} illustrates sensing/transition corruption \cite{chattopadhyay2021robustnav}, locomotion mismatch in actuation and whole-body dynamics \cite{cha2025sim}, and deformable-object errors caused by simplified materials, friction, self-contact, and deformation physics \cite{synreal2026,style3d2026deformable}.

In practice, failures in navigation can arise from perception, geometry, and collision mismatches; legged or humanoid systems additionally introduce locomotion and whole-body control mismatches; and manipulation depends on robot dynamics, contact modeling, material properties, and, in deformable-object settings, soft-body physics. Moreover, the same high-level metric can hide different simulator errors. For example, two navigation agents may have similar SPL but different collision responses or clearance margins, two locomotion policies may have similar simulated return but different hardware drift or fall rates, and two manipulation policies may have similar simulated success but differ in contact force, slip margin, deformation behavior, or robustness to material variation.  Table~\ref{tab:error_analysis} further lists representative sources of simulator-induced mismatches, their observable failures, and practical checks or mitigations. These errors can arise from various factors, including observation generation, transition dynamics, robot control, environment modeling, and benchmark evaluation.

\begin{table*}[htbp]
\caption{Concise taxonomy of simulator-induced failure cases. The rows provide fine-grained examples of mismatch sources underlying the representative failure cases illustrated in Fig.~\ref{fig:error_schematics}.}
\label{tab:error_analysis}
\centering
\footnotesize
\setlength{\tabcolsep}{3pt}
\renewcommand{\arraystretch}{1.08}
\begin{tabularx}{\textwidth}{@{}L{3.2cm} Y Y@{}}
\toprule
\textbf{Mismatch source}
& \textbf{Failure signature}
& \textbf{Check / mitigation} \\
\midrule

Geometry / collision \cite{kadian2020sim2real}
& Wall sliding, corner cutting, or high SPL with unsafe contacts.
& Log contacts, clearance, and path deviation; test sim--real rank agreement; tune navmesh or use stop-on-contact evaluation. \\

Observation / sensing \cite{chattopadhyay2021robustnav,pumacay2024colosseum}
& Goal or object grounding fails under lighting, texture, depth, FOV, blur, or latency shifts.
& Run factorized corruption tests; calibrate RGB-D sensors; model noise/latency; use augmentation or domain randomization. \\

Dynamics, actuation, and timing \cite{tan2018simtoreal,kataoka2023bimanual}
& Slip, falls, oscillation, overshoot, desynchronization, or success only under idealized APIs.
& Compare commands with measured states and timestamps; apply system identification, dynamics randomization, delay modeling, and hardware-matched controllers. \\

Rigid contact / material \cite{collins2019quantifying}
& Grasp slip, penetration, bounce/stick behavior, wrong object rotation, or insertion jamming.
& Inspect contacts, penetration, force/torque, sensitivity to friction, mass, and timestep; calibrate meshes, materials, and solvers. \\

Deformable, fragile, and tactile objects \cite{blanco2024benchmarking,narang2021tactile}
& Cloth folds, rope knots, fracture thresholds, or tactile images/forces do not transfer.
& Compare deformation, self-contact, force, and tactile; identify material parameters and calibrate tactile or soft-body models. \\

Evaluation artifact \cite{kadian2020sim2real,pumacay2024colosseum}
& Similar SR/SPL or success rate hides different collisions, forces, slip margins, or robustness.
& Report metrics: contact rate, clearance, force/slip margin, perturbation robustness, and sim--real predictivity. \\

\bottomrule
\end{tabularx}
\end{table*}

\subsection{Future Directions}

\noindent \textbf{Continual Learning.}
Continual learning is critical for embodied agents to adapt to dynamic environments while retaining prior knowledge, particularly in VLN, where catastrophic forgetting poses a challenge \cite{li2024visionlanguagenavigationcontinuallearning}. Emerging approaches involve regularization methods such as Elastic Weight Consolidation (EWC) \cite{Kirkpatrick_2017}, replay-based methods, Meta-learning \cite{finn2017modelagnosticmetalearningfastadaptation}, and memory architectures such as Titans neural long-term memory modules \cite{behrouz2024titanslearningmemorizetest}. Recent studies further advance this area: NeSyC \cite{choi2025nesyc} introduces a neuro-symbolic continual learner that integrates neural and symbolic reasoning for complex tasks in open domains, while Zheng et al. \cite{zheng2025lifelong} provide a roadmap for lifelong learning in LLM-based agents, emphasizing perception, memory, and action modules to enhance adaptability and mitigate forgetting.

\noindent \textbf{Neural ODEs.}
Embodied AI tasks (\eg pouring liquids) require continuous dynamics modeling, challenging for discrete methods. Neural ODEs \cite{chen2019neuralordinarydifferentialequations} enable continuous state evolution, improving trajectory prediction and control under variables such as object masses. Liquid networks \cite{hasani2020liquidtimeconstantnetworks} process irregular inputs (\eg cameras) for real-time adaptation. Though promising for precise manipulation, empirical validation remains essential.

\noindent \textbf{Evaluation Metrics.}
Current evaluation metrics are overly goal-oriented (\eg success rate, path length). We advocate for procedural quality metrics inspired by human task execution, such as \textbf{Energy efficiency} (Minimize energy expenditure) and \textbf{Smoothness} (Quantify abrupt changes in trajectory). Recent benchmarks, such as the exploration-aware EQA framework by Jiang et al. \cite{EXPRESSBench}, expand this scope by emphasizing exploration in task evaluation, providing a more comprehensive assessment of Embodied AI performance.

\noindent \textbf{Safe and Trustworthy Evaluation.} 
A key limitation of current simulator-based benchmarks is that task success in simulation does not necessarily imply safe or natural real-world execution. A navigation agent may obtain high SPL by exploiting collision-response artifacts, while a manipulation policy may complete a task despite unsafe contact, unstable release, excessive force, cross-contamination, or incorrect temporal ordering. Recent evidence from SAFEMANIP \cite{huang2026safemanip} shows that task-success gains in robotic manipulation do not reliably translate into temporal safety gains, and that many successful rollouts are completed but unsafe. Future benchmarks should therefore move beyond success rates and report temporal safety violations, motion naturalness, and sim-to-real predictivity. In practice, this requires property-level runtime monitors, safety-margin measurements, and real-world or hardware-in-the-loop validation to determine whether simulator performance is relevant for deployment.
\section{Conclusions} \label{sec:conclusion}
In this survey, we have reviewed robotic navigation and manipulation in Embodied AI, with a focus on physics simulators and sim-to-real transfer. We examined simulator properties that are often underemphasized in prior surveys and discussed how they affect learning, evaluation, and real-world deployment. We also consolidated recent advances in tasks, benchmark datasets, evaluation metrics, simulation platforms, and representative methods for navigation and manipulation, ranging from classical reinforcement and imitation learning to world models, diffusion policies, foundation models, and vision-language-action models. Our analysis shows that simulator selection should be driven by task-specific physical requirements and hardware constraints, since different settings impose different sensing, dynamics, and compute requirements. Finally, by summarizing simulator-induced failure modes and future research directions, this survey provides a practical reference for selecting suitable tools and designing simulation-based studies that better anticipate real-world embodied-agent performance.

\bibliographystyle{ACM-Reference-Format}
\bibliography{references}

\newpage

\end{document}